\UseRawInputEncoding
\documentclass{article}
\usepackage{float}
\usepackage[margin=1.0in]{geometry}
\usepackage{abstract}
\usepackage{afterpage}
\usepackage[backend=bibtex,style=nature,natbib=true,maxbibnames=99,minalphanames=3]{biblatex}
\usepackage{hyperref}

\usepackage{caption}
\usepackage{lmodern}
\usepackage{threeparttable}
\usepackage[dvipsnames,svgnames, table,xcdraw]{xcolor}
\usepackage{relsize}
\usepackage{tcolorbox}
\usepackage{listings}
\usepackage{subcaption}
\usepackage{algorithm}
\usepackage{algorithmic}
\usepackage{comment}
\usepackage{graphicx}
\usepackage{booktabs}
\usepackage{multirow}
\usepackage{todonotes}
\usepackage{enumitem}
\usepackage{url}
\usepackage{mathpazo}
\usepackage{amsmath,amssymb,amsthm,amsfonts,bm}
\usepackage[toc,page,header]{appendix}
\usepackage{fancyhdr}
\usepackage[T1]{fontenc}

\usepackage{auth_detailed}

\counterwithout{figure}{section}
\counterwithout{table}{section}
\definecolor{darkgoldenrod}{rgb}{0.72, 0.53, 0.04}
\definecolor{backgroundcolor}{RGB}{250, 250, 252}   
\definecolor{keywordcolor}{RGB}{30, 0, 178}       
\definecolor{stringcolor}{RGB}{204, 0, 102}        
\definecolor{numbercolor}{RGB}{0, 128, 128}        
\definecolor{emphcolor}{RGB}{30, 0, 178}            
\definecolor{commentcolor}{RGB}{0, 128, 0}       
\definecolor{basiccodecolor}{RGB}{61, 61, 61}       

\lstdefinestyle{customstyle}{
    backgroundcolor=\color{backgroundcolor},   
    commentstyle=\color{commentcolor},
    keywordstyle=\color{keywordcolor},
    numberstyle=\color{numbercolor},
    stringstyle=\color{stringcolor},
    basicstyle=\color{basiccodecolor}\ttfamily\footnotesize,
    breakatwhitespace=false,         
    breaklines=true,                 
    captionpos=b,                    
    keepspaces=true,                 
    numbers=left,     
    basicstyle=\color{basiccodecolor}\ttfamily\footnotesize,
    numbersep=5pt,             
    xleftmargin=2em,
    xrightmargin=2em,
    showspaces=false,                
    showstringspaces=false,
    showtabs=false,                  
    tabsize=1,
    frame=single,
    framesep=5pt,
    framexleftmargin=1.5em,
    framexrightmargin=1.5em,
    framextopmargin=1pt,
    framexbottommargin=1pt,
    aboveskip=10pt,
    belowskip=10pt,
    breaklines=true,
    breakautoindent=true,
    emph={},
    emphstyle={\color{emphcolor}},
    extendedchars=true,
}

\usepackage[normalem]{ulem}
\usepackage{xcolor}

\newcommand{\scrcmd}{\textsuperscript}
\newcommand{\scr}[1]{\scrcmd{#1}}

\usepackage{fontawesome5}
\usepackage{helvet}
\DeclareRobustCommand{\corrAuthor}{\textsuperscript{\faEnvelope[regular]}}

\usepackage{soul}
\sethlcolor{yellow}
\usepackage{hyperref}
\usepackage{xcolor}
\hypersetup{breaklinks=true}

\hypersetup{
  colorlinks=true,
  urlcolor=blue,
  citecolor=ForestGreen,
  linkcolor=blue
}
\newcommand{\arialtitle}[1]{{\fontfamily{phv}\selectfont #1}}

\newif\ifpreprint
\preprinttrue    

\title{{\fontsize{15pt}{18pt}\selectfont \textbf{\arialtitle{
\ifpreprint
\raisebox{-0.25\height}{\includegraphics[height=1.1em]{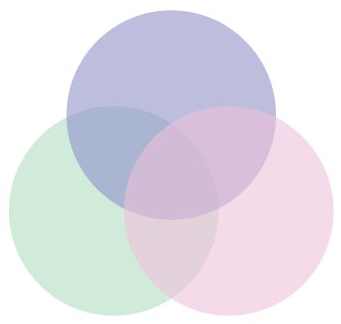}}\hspace{0.3em}
\fi
Linking spatial biology and clinical histology via Haiku}}}}

\author{\name Yan Cui$^{1,2,*}$,
\name Jacob S. Leiby$^{1,*}$,
\name Wenhui Lei$^{1}$,
\name Dokyoon Kim$^{3}$,
\name Yanxiang Deng$^{1,2}$,
\name Aaron T. Mayer$^{4}$,\\
\name Zhenqin Wu$^{4,\corrAuthor}$,
\name Alexandro E. Trevino$^{4,\corrAuthor}$,
\name Zhi Huang$^{1,3,\corrAuthor}$\\\newline
$^{1}$ Department of Pathology and Laboratory Medicine, University of Pennsylvania, Philadelphia, PA, USA\\
$^{2}$ Department of Bioengineering, University of Pennsylvania, Philadelphia, PA, USA\\
$^{3}$ Department of Biostatistics,
Epidemiology \texorpdfstring{\&}{&} Informatics, University of Pennsylvania, Philadelphia, PA, USA\\
$^{4}$ Enable Medicine, Menlo Park, CA, USA\\
*~~~Equal contribution\\
\corrAuthor~Correspondence:\\
Zhi Huang (\href{mailto:zhi.huang@pennmedicine.upenn.edu}{zhi.huang@pennmedicine.upenn.edu})\\
Alexandro E. Trevino (\href{mailto:alex@enablemedicine.com}{alex@enablemedicine.com})\\
Zhenqin Wu (\href{mailto:zhenqin@enablemedicine.com}{zhenqin@enablemedicine.com})\\
}

\begin{document}



\maketitle

\ifpreprint
\fancyhead[L]{\raisebox{-0.1\height}{\includegraphics[height=1.2em] {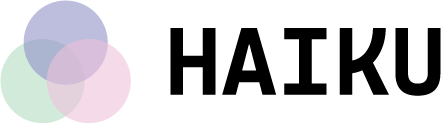}}}
\fancyhead[R]{Linking H\&E, mIF, and text in a unified framework}
\setlength{\headheight}{13pt}
\pagestyle{fancy}
\fi

\renewcommand{\abstractnamefont}{\normalfont\normalsize\bfseries}
\renewcommand{\abstracttextfont}{\normalfont\normalsize}

\renewcommand{\abstractname}{Abstract}
\begin{abstract}
\abstracttextfont

Integrating molecular, morphological, and clinical data is essential for basic and translational biomedical research, yet systematic frameworks for jointly modeling these modalities remain limited. Here we present Haiku, a tri-modal contrastive learning model trained on multiplexed immunofluorescence (mIF). It comprises 26.7 million spatial proteomics patches from 3,218 tissue sections across 1,606 patients spanning 11 organ type, with matched hematoxylin and eosin (H\&E) histology and clinical metadata aligned in a shared embedding space. Haiku enables three-way cross-modal retrieval, improves a variety of downstream classification and clinical prediction tasks over unimodal baselines, and supports zero-shot biomarker inference through fusion retrieval conditioned on clinical metadata-only text descriptions. Across tasks, Haiku outperforms competing approaches, achieving cross-modal retrieval (Recall@50 up to 0.611 versus near-zero baseline), survival prediction (C-index 0.737, +7.91\% relative improvement), and zero-shot biomarker inference (mean Pearson correlation 0.718 across 52 biomarkers). Furthermore, we introduce a counterfactual prediction framework in which modifying only clinical metadata while fixing tissue morphology surfaces niche-specific molecular shifts associated with breast cancer stage progression and lung cancer survival outcomes. In a lung adenocarcinoma case study, the counterfactual analysis recovers niche-specific shifts characterized by increased CD8 and granzyme B, reduced PD-L1, and decreased Ki67, broadly consistent with patterns reported in the literature for favorable outcomes. We present these counterfactual results as exploratory, hypothesis-generating signals rather than mechanistic claims. These capabilities demonstrate that tri-modal alignment via Haiku enables integrative analysis of spatial biology, bridging molecular measurements with clinical context to support biological exploration and downstream investigation.

\end{abstract}

\section{Introduction}

Modern biological and translational research increasingly relies on integrating molecular, imaging, and clinical data to understand disease mechanisms and guide therapeutic decisions. The recent emergence of spatial omics and large-scale histopathology has made this integration both more urgent and more tractable, providing complementary, richly structured views of tissue biology. The rapid accumulation of these high-dimensional, multimodal datasets presents a fundamental computational challenge: learning the multi-directional relationships between data modalities. These relationships likely encode biological and clinical insights not accessible through the analysis of any single data type alone.

Progress on deep learning within individual modalities has been substantial and motivates this challenge directly. In histopathology, foundation models trained on hematoxylin and eosin (H\&E) stained slides have enabled representation learning~\cite{Chen2024-sl, Ding2025-mv}, biomarker discovery~\cite{Li2025-kj, Swanson2025-cd, Huang2025-kh}, and prediction of patient prognosis, treatment outcomes, and clinical markers~\cite{Campanella2019-zp, Ahn2024-yu, Yao2020-xz}. In parallel, spatial proteomics has allowed simultaneous quantification and localization of 50 or more protein antigens in a single tissue section~\cite{Black2021-dh}, yielding molecular insights into the tumor microenvironment~\cite{Elhanani2023-rs, Shaitelman2024-be, Foster2022-tn, Chen2025-ji}. Dedicated AI encoders have been developed for these omics modalities as well~\cite{Wenckstern2025-ve, shaban2025foundation, liu2025modeling}. In a complementary direction, structured clinical metadata and free-text reports have likewise been incorporated into pathology models, primarily through vision--language pretraining or text-conditioned annotation~\cite{Huang2023-di, Lu2024-dz, Xiang2025-vd}, providing a third stream of tissue-level semantic context. Yet each modality captures only one perspective on a given tissue. Additional biological insights may remain latent in the interactions between modalities. 

Critically, these modalities are not independent: H\&E morphology, spatial protein expression, and clinical or semantic context represent complementary views of the same underlying tissue biology. Recent work has begun to exploit parts of these relationships by predicting spatial proteomics directly from H\&E images~\cite{Wu2025-nj, Valanarasu2025-ek, Li2026-kw}, aligning histopathology with text or clinical metadata~\cite{Huang2023-di, Chen2025-ee, Lu2024-dz}, and using semantic descriptors as bridges between molecular data and biological interpretation~\cite{Xiang2025-vd, Wang2024-hs, Yiyao2025-fi}. However, these approaches remain largely pairwise, task-specific, or focused on modality imputation rather than joint multimodal representation learning. To our knowledge, frameworks that jointly model H\&E histology, spatial proteomics, and clinical or semantic context within a unified, bidirectional representation remain limited. 

As a result, current models fall short of fully exploiting multi-directional relationships across multimodal biomedical data for deeper reasoning and discovery. Jointly representing spatial biology, tissue morphology, and clinical context in a unified framework could lead to more relevant and actionable biomedical insights. To fully realize the potential of modern biomedical data, there is a pressing need for models that not only integrate diverse modalities into a shared representation, but also enable systematic exploration of these representations to uncover latent biological signals and mechanisms and to generate new hypotheses. Motivated by these unmet needs and clear gaps, here we propose \emph{Haiku}, a pretrained tri-modal contrastive learning AI model that jointly integrates H\&E images, multiplexed immunofluorescence (mIF) images, and textual information, including patch-level descriptors and tissue-level clinical descriptions. Pretrained on 26,669,005 proteomics image patches spanning 120 unique biomarkers across 3,218 paired tissue sections from 1,606 patients across 11 organ types and 11 diseases (\textbf{Supplementary Figure~\ref{supplementary:1}}), \emph{Haiku} encourages mutual alignment and cycle consistency among all three modalities. 

Through large-scale multimodal pretraining, \emph{Haiku} unifies diverse data modalities within a shared representation, enabling coherent cross-modal alignment, retrieval, and integration of molecular, morphological, and clinical information; this unified framework, in turn, enables the discovery of latent biological knowledge and the potential generation of new insights and hypotheses. The resulting embedding space supports robust retrieval across modalities, achieving Recall@$50$ of 0.604 for mIF-to-H\&E, 0.611 for H\&E-to-mIF, and 0.169 for Text-to-mIF on held-out data, while also improving downstream clinical prediction tasks, including a mean C-index of 0.737 for colorectal cancer survival prediction and AUPRC values of 0.660 and 0.775 for melanoma and colorectal cancer treatment-response prediction. Building on this unified representation, we further introduce zero-shot fusion retrieval for biomarker inference, which reaches a mean Pearson correlation of 0.718 across 52 biomarkers, and a counterfactual prediction framework that surfaces niche-specific molecular shifts associated with cancer progression and survival outcomes, presented as exploratory, hypothesis-generating analyses.

In summary, \emph{Haiku} provides a unified framework that integrates heterogeneous biomedical data, enables improved predictive and retrieval performance, and supports their systematic exploration for knowledge discovery. Most importantly, by bridging spatial biology with clinical and semantic representations, \emph{Haiku} establishes a new paradigm for discovering latent molecular and microenvironmental programs and generating new biological hypotheses. Source code and model checkpoint are available at \url{https://github.com/zhihuanglab/Haiku}

\section{Results}
\label{sec:benchmarks}

\begin{figure}[hbtp]
    \centering
    \includegraphics[width=\textwidth]{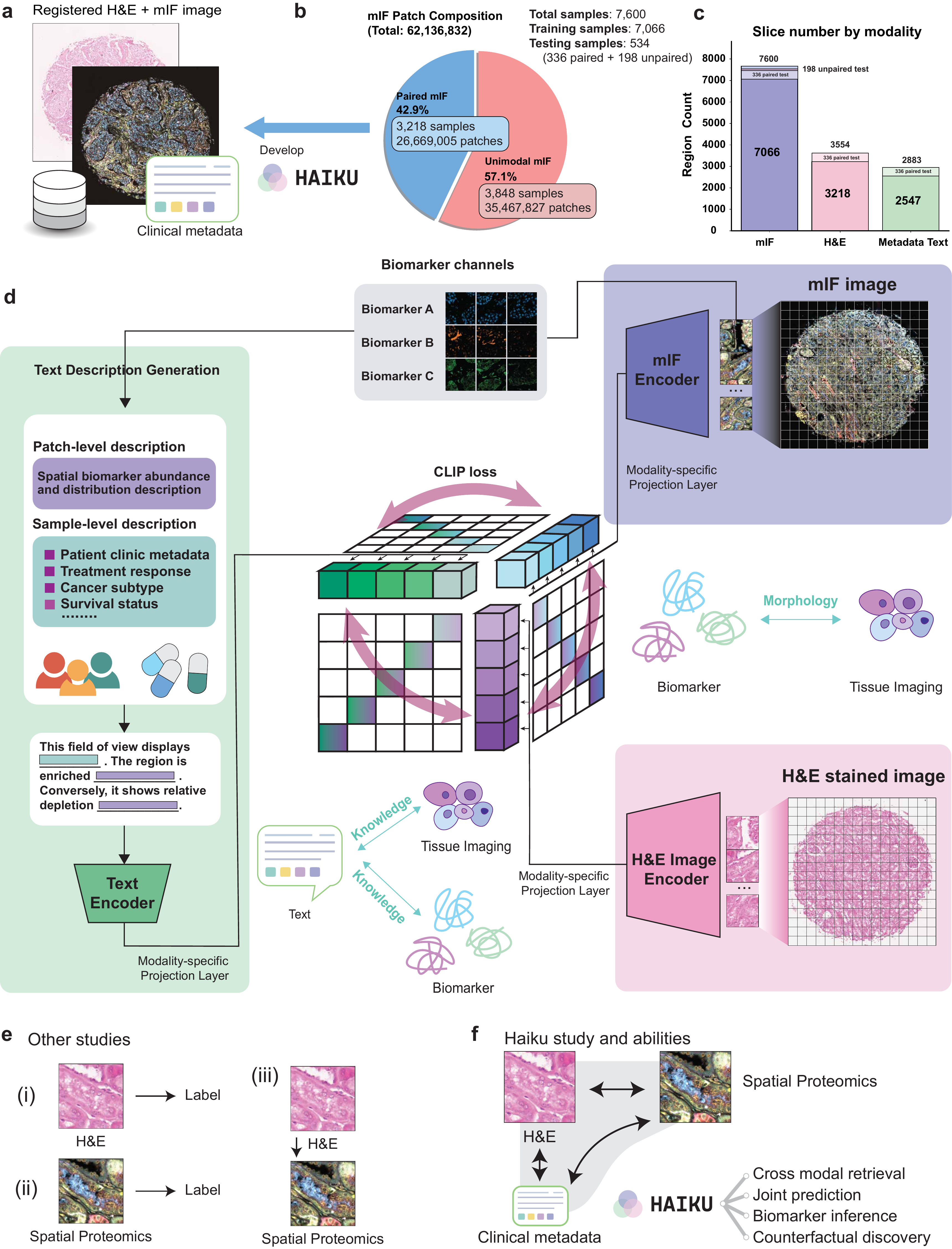}
    \captionsetup{
      format=plain,
      justification=raggedright,
      singlelinecheck=false,
      margin=0pt,
      width=\dimexpr\textwidth\relax
    }
    \caption[\textbf{Dataset composition and overview of the Haiku framework.}]%
    {\textbf{Dataset composition and overview of the Haiku framework.} See next page for caption.}
    \label{fig:figure1}
\end{figure}

\begin{figure}[hbtp]
    \ContinuedFloat
    \captionsetup{list=no}
    \caption{
    \small
    \textbf{Dataset composition and overview of the Haiku framework.}
    \textbf{a}, Representative example of a co-registered H\&E and mIF tissue slice together with its associated patient-level clinical metadata, illustrating the three paired data streams that serve as input to Haiku.
    \textbf{b}, Composition of the curated mIF training-patch corpus (total: 62{,}136{,}832 patches from 7{,}066 training slices). Paired H\&E--mIF slices (42.9\%; 3{,}218 samples, 26{,}669{,}005 patches) are used for tri-modal contrastive alignment, whereas unimodal mIF-only slices (57.1\%; 3{,}848 samples, 35{,}467{,}827 patches) are used exclusively for pretraining the mIF encoder. The full mIF corpus comprises 7{,}600 slices, of which 7{,}066 are assigned to training and 534 are reserved for testing (336 paired held-out $+$ 198 unpaired mIF-only held-out).
    \textbf{c}, Slice counts for each of the three modalities (mIF, H\&E, and metadata text), shown as stacked bars in which the lower (saturated) segment denotes training slices and the upper (desaturated) segments denote held-out test slices. The mIF modality contributes 7{,}600 slices in total (7{,}066 training $+$ 336 paired held-out $+$ 198 unpaired held-out); H\&E contributes 3{,}554 (3{,}218 training $+$ 336 paired held-out); and metadata text contributes 2{,}883 (2{,}547 training $+$ 336 paired held-out), highlighting the multimodal coverage of the training and evaluation corpus.
    \textbf{d}, Schematic of the Haiku architecture. Three modality-specific encoders process H\&E images, mIF images, and text descriptions, respectively. The text descriptions combine patch-level spatial biomarker abundance and distribution patterns with sample-level patient clinical metadata. Embeddings from all three modalities are passed through modality-specific projection layers and aligned in a shared latent space via pairwise CLIP-style contrastive losses, yielding a tri-modal representation that links morphology, biomarker, and clinical knowledge.
    \textbf{e}, Scope of existing modeling paradigms in computational pathology, which typically operate in a unimodal-to-label fashion: (i) H\&E-to-label, (ii) spatial proteomics-to-label, and (iii) H\&E-to-spatial-proteomics translation, each targeting a single task or modality pair.
    \textbf{f}, Overview of the study scope and capabilities enabled by Haiku. By jointly aligning H\&E, spatial proteomics, and clinical metadata within a single embedding space, Haiku supports cross-modal retrieval, joint prediction, biomarker inference, and counterfactual discovery as downstream applications of a single pretrained model.
    }
\end{figure}

\subsection{Training a unified multimodal foundation model ``Haiku'' for joint representation learning of H\&E, mIF, and text}

Haiku is built on a multi-center, multi-disease cohort, comprising tri-modal tissue samples in which co-registered H\&E and multiplexed immunofluorescence (mIF) slices are paired with patient-level clinical metadata (\textbf{Figure~\ref{fig:figure1}a}). The full mIF corpus comprises 7{,}600 tissue slices, of which 7{,}066 are used for training (62{,}136{,}832 patches) and 534 are reserved for held-out testing (336 paired $+$ 198 unpaired); within the training pool, 3{,}218 slices (contributing $42.9\%$ of training patches; 26{,}669{,}005 patches) carry paired H\&E and metadata and are used for tri-modal contrastive alignment, while the remaining 3{,}848 slices (contributing $57.1\%$ of training patches; 35{,}467{,}827 patches) are mIF-only and used exclusively to pretrain the mIF encoder (\textbf{Figure~\ref{fig:figure1}b}). At the patient level, the cohort comprises $1{,}848$ patients in total, partitioned into $1{,}606$ ($86.9\%$) for training and $242$ ($13.1\%$) for held-out testing, with the split performed at the patient level to prevent any patient-level leakage between training and evaluation (\textbf{Supplementary Figure~\ref{supplementary:2}}). Across modalities, the cohort provides 7{,}600 mIF, 3{,}554 H\&E, and 2{,}883 metadata-text slices in total (training $+$ held-out test), ensuring broad coverage for both unimodal pretraining and tri-modal alignment (\textbf{Figure~\ref{fig:figure1}c}).

Based on this data collection, we develop Haiku, a foundation model built upon well-established cross-modality alignment paradigms in general multimodal representation learning via contrastive learning. Haiku extends these frameworks to a tri-modal setting by jointly modeling H\&E images, spatial proteomics, and textual information. For the input data, Haiku first partitions well-registered, paired H\&E and mIF images into $256 \times 256$ pixel patches. Specifically, for each tissue slice, the H\&E image and the corresponding mIF image are exactly co-registered, enabling the same patching procedure to be applied to both modalities and yielding spatially paired H\&E patches and multi-channel mIF patches (\textbf{Figure~\ref{fig:figure1}d}). 

In addition to image-based modalities, we generate two textual descriptions for local patch-level information and global slice-level clinical context. Global clinical descriptions are constructed from paired patient metadata, such as treatment response information, prognosis information, partial pathological descriptions, tissue section annotations, tumor type, and tissue type. These metadata are formatted into structured textual descriptions and assigned to all patches belonging to the corresponding tissue region. For patch-level descriptions, we compute intra-slice z-scores for each biomarker channel to categorize biomarker abundance as low or high, and further augment these descriptions with spatial distribution patterns to capture spatial contextual information~\cite{Yiyao2025-fi}. By combining slice-level clinical descriptions with patch-level biomarker and spatial information, we generate a final textual description for each image patch (\textbf{Figure~\ref{fig:figure1}d}; Methods~\ref{sec:textdescript}). For slices without patient-level clinical metadata, the text modality consists solely of the patch-level biomarker abundance and spatial-distribution description, omitting the sample-level clinical context. This process yields well-aligned text, mIF, and H\&E modalities for subsequent model training.

In contrastive learning--based multimodal alignment, prior work commonly leverages large-scale pretrained unimodal or multimodal encoders to provide good initialization for downstream alignment. Existing computational pathology efforts have largely focused on unimodal-to-label prediction or on pairwise translation between H\&E and spatial proteomics (\textbf{Figure~\ref{fig:figure1}e}), leaving tri-modal integration with clinical text under-explored. Following the contrastive-alignment paradigm, we adopt the pretrained MUSK~\cite{Xiang2025-vd} encoder for histology image representation and incorporate a pretrained biomedical BERT encoder (BiomedBERT~\cite{Gu2021-biomedbert}) to represent clinical text. We then train an mIF encoder from scratch on our in-house mIF dataset (paired + unpaired; 7,066 slices, 62,136,832 patches; \textbf{Figure~\ref{fig:figure1}b}) using the VirTues~\cite{Wenckstern2025-ve} architecture, yielding a high-capacity pretrained encoder for spatial proteomics data. To ensure a fair comparison, the VirTues baseline reported throughout this paper is pretrained on the same dataset. For each modality, we construct a modality-specific projection head that maps modality-specific embeddings into a shared latent space. These projection heads are trained using a contrastive loss~\cite{Radford2021-ib} (Methods~\ref{sec:training}), in which matched H\&E patches, mIF patches, and corresponding textual descriptions are treated as positive samples, while all other combinations are treated as negatives. This objective encourages embeddings originating from the same patch but represented in different modalities to align closely in the shared embedding space (\textbf{Figure~\ref{fig:figure1}d}).

We then train Haiku on a dataset containing 3,218 paired tissue slices with 26{,}669{,}005 patches, each with all three paired modalities (\textbf{Figure~\ref{fig:figure1}b,c}). The pretraining corpus spans diverse organ types and disease categories, including breast, lung, colon, kidney, and liver tissues across multiple cancer types and normal tissues (\textbf{Supplementary Figure~\ref{supplementary:1}}). After training, the pretrained encoders can be directly used to extract features from each modality, enabling efficient and scalable cross-modality retrieval and improving performance on diverse patch- and slide-level supervised downstream tasks (\textbf{Figure~\ref{fig:figure1}f}).

\subsection{Haiku enables generalized cross-modality retrieval among H\&E, mIF, and text modalities}

In addition to the training dataset, we reserve a set of 336 held-out paired slices for generalization evaluation. We first assess a fundamental and clinically relevant application: cross-modality retrieval (\textbf{Figure~\ref{fig:figure2}a}). This task is both research and clinically relevant because an accurate retrieval is a direct signal to indicate that these data modalities are well-aligned during learning. Specifically, we evaluate patch-level retrieval tasks in which a query patch of one modality (H\&E, mIF, or text) is used to retrieve patches from a different target modality (H\&E or mIF). Exact matches are defined as retrievals of target patches from the same spatial location. Retrieval performance is evaluated using top-$k$ recall with $k \in \{1, 5, 10, 20, 50\}$ (See Methods~\ref{sec:retrieval} for more details).

Importantly, retrieval is conducted across the entire cross-dataset reference collection, rather than being restricted to patches from the same tissue slice or study. In practice, each query is evaluated against 336 held-out slices spanning multiple datasets. Compared with conventional intra-dataset retrieval settings, this scenario is substantially more challenging but also considerably more realistic and clinically applicable. In this setting, users can directly apply pretrained Haiku to their own H\&E patches or text descriptions and retrieve relevant mIF patches from a large-scale reference atlas. Conversely, when users profile mIF data and require corresponding H\&E patches, the same framework enables retrieval across large-scale H\&E reference collections.

We first present qualitative retrieval examples to illustrate the effectiveness of Haiku's cross-modal alignment (\textbf{Figure~\ref{fig:figure2}a}). We begin with the example of Text-to-mIF retrieval (\textbf{Figure~\ref{fig:figure2}b}). In this setting, the query text describes a breast cancer sample with associated clinical metadata and spatial biomarker patterns. We visualize the top-ranked retrieval candidates alongside biomarker channel comparisons with the ground-truth mIF patch, focusing on biomarkers explicitly mentioned in the text description. The top-$1$ retrieved candidate exactly matches the ground truth. Importantly, other retrieved candidates faithfully reflect the spatial biomarker descriptions (\textbf{Figure~\ref{fig:figure2}b}) in the text. For example, biomarkers described as enriched, such as GranzymeB, CD11c, and PanCK, display consistent, high expression across all top-ranked candidates. Meanwhile, biomarkers described as having lower abundance, including Ki67 and IFN$\gamma$, are also reflected consistently across candidates. These results demonstrate not only the effectiveness of Haiku's Text-to-mIF retrieval but also the reasoning fidelity of our text generation pipeline for spatial biomarker descriptions.

We next present an H\&E-to-mIF retrieval example (\textbf{Figure~\ref{fig:figure2}c}). Given an H\&E patch as a query, Haiku retrieves the corresponding mIF patch at rank~1. Moreover, the remaining top-ranked candidates exhibit highly similar mIF spatial distributions and consistent H\&E morphological patterns, indicating that Haiku embeddings jointly capture tissue morphology and spatially resolved molecular organization. Visualization of individual biomarker channels (\textbf{Figure~\ref{fig:figure2}c}) further confirms that the model accurately encodes per-biomarker spatial patterns using H\&E input alone: top-ranked retrieved patches consistently display biomarker-specific spatial distributions that closely match the ground truth.

To demonstrate Haiku's capabilities, we conduct detailed quantitative benchmarking comparisons. Since Haiku is, to our knowledge, the first model to attempt patch-level cross-modality retrieval between mIF, H\&E, and text jointly, no existing baseline methods directly apply as comparisons. To show that this task is non-trivial and to provide a reference baseline, we construct a naive approach by stacking mIF channels into a three-channel RGB-like format and feeding them into the same fine-tuned H\&E encoder used by Haiku prior to the projection layers. Haiku achieves strong performance, reaching high Recall@$1$ accuracy in image-based retrieval tasks (H\&E-to-mIF and mIF-to-H\&E) and showing consistent improvement as the top-$k$ parameter increases. Notably, Haiku attains a Recall@$50$ of $0.604$ in the mIF-to-H\&E retrieval task (\textbf{Figure~\ref{fig:figure2}d}) and $0.611$ in the H\&E-to-mIF retrieval task (\textbf{Figure~\ref{fig:figure2}e}), whereas the naive baseline yields near-zero performance across all settings. These results demonstrate that image-based cross-modality retrieval can achieve ready-to-use performance through our contrastive learning strategy. More importantly, they provide strong evidence that higher-order relationships exist between the spatially resolved biomedical information encoded in mIF and the morphology-dominated H\&E modality, and that these relationships can be aligned within a shared low-dimensional latent representation. Text-based mIF retrieval (Text-to-mIF) achieves relatively lower performance due to the inherent information gap between modalities; nevertheless, despite the challenging cross-dataset reference setting, it still reaches a $0.169$ Recall@$50$, indicating meaningful alignment in the latent space (\textbf{Figure~\ref{fig:figure2}f}).

Beyond retrieval, we evaluate zero-shot patch-level annotation across image modalities using 1-nearest-neighbor classification (\textbf{Figure~\ref{fig:figure2}g--i}). Annotation labels include organ type, tumor grade, and tissue type. Given a query patch from one modality, its label is assigned based on the labels of the retrieved top-$1$ patches from the target modality. This task assesses whether Haiku captures consistent tissue-context information and ensures that retrieved patches share coherent pathological and anatomical characteristics. Across all evaluated tasks, Haiku consistently outperforms MUSK and majority-voting baselines, achieving a $0.842$ macro-averaged F1 score on organ-type classification in the mIF-to-H\&E setting (\textbf{Figure~\ref{fig:figure2}g}), further demonstrating the high quality of the learned embeddings.

We further perform CLIP-style zero-shot classification~\cite{Huang2023-di} by querying mIF patches against text prompts such as ``A mIF image of \_'' (\textbf{Figure~\ref{fig:figure2}j}). We evaluate two tasks: organ-type classification (10 categories) and disease classification (11 categories), using the full category sets observed in the paired held-out test dataset (see \textbf{Supplementary Figure~\ref{supplementary:3}} for the complete category list and per-class proportions). Despite the inherent difficulty of zero-shot classification on mIF patches using only natural-language prompts, and the substantial class imbalance across these 10- and 11-way tasks, Haiku substantially outperforms a random-paired baseline on both, reaching a macro-averaged F1 of $0.179$ for organ type (vs.\ $0.067$ random baseline) and $0.182$ for disease (vs.\ $0.059$ random baseline), with the difference statistically significant in both settings (two-sided Wilcoxon rank-sum (Mann--Whitney $U$) test; $P < 0.001$) (\textbf{Figure~\ref{fig:figure2}k}). These results demonstrate stable zero-shot capability and further confirm the strong cross-modal alignment achieved by the Haiku embedding space.

\begin{figure}[hbtp]
    \centering
    \includegraphics[width=\textwidth]{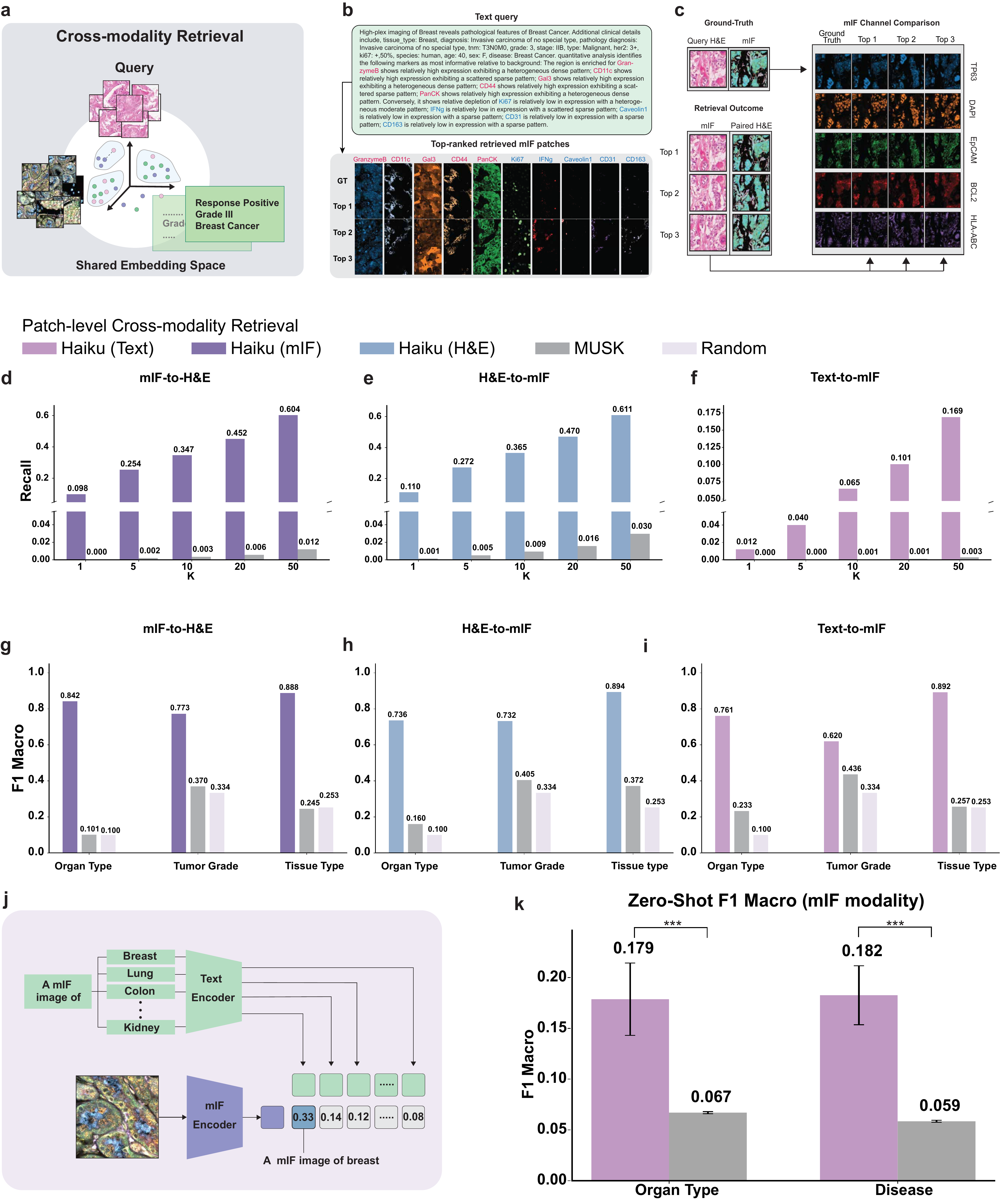}
    \captionsetup{
      format=plain,
      justification=raggedright,
      singlelinecheck=false,
      margin=0pt,
      width=\dimexpr\textwidth\relax
    }
    \caption[\textbf{Cross-modality alignment, retrieval, and zero-shot evaluation.}]%
    {\textbf{Cross-modality alignment, retrieval, and zero-shot evaluation.} See next page for caption.}
    \label{fig:figure2}
\end{figure}

\begin{figure}[hbtp]
    \ContinuedFloat
    \captionsetup{list=no}
    \caption{
    \small
    \textbf{Cross-modality alignment, retrieval, and zero-shot evaluation.}
    \textbf{a}, Conceptual schematic of patch-level cross-modality retrieval. Queries from any of the three modalities (H\&E, mIF, or text describing clinical context and biomarker patterns) are projected into the shared Haiku embedding space, enabling retrieval of semantically matched patches from the target modality.
    \textbf{b}, Text-to-mIF retrieval example. The input text describes a breast cancer sample with color-coded biomarker abundance and spatial patterns; ground-truth and top-3 retrieved mIF patches are shown for each explicitly mentioned biomarker channel, illustrating that retrieved patches faithfully reflect the semantic content of the query.
    \textbf{c}, H\&E-to-mIF retrieval example. For a query H\&E patch, the ground-truth mIF and the top-3 retrieved mIF patches are shown alongside their paired H\&E; per-biomarker channel comparisons (TP63, DAPI, EpCAM, BCL2, HLA-ABC) highlight consistency between retrieved and ground-truth spatial signals.
    \textbf{d--f}, Quantitative Recall@$K$ benchmarking ($K=1,5,10,20,50$) for cross-modality retrieval across the full cross-dataset reference pool: (\textbf{d}) mIF-to-H\&E, (\textbf{e}) H\&E-to-mIF, and (\textbf{f}) Text-to-mIF. Haiku is compared with the MUSK baseline (naive RGB-stacking of mIF channels into the same fine-tuned H\&E encoder) and a random baseline.
    \textbf{g--i}, Zero-shot 1-nearest-neighbor patch-level annotation using retrieved patches, reported as macro-averaged F1 across three label types (organ type, tumor grade, and tissue type): (\textbf{g}) mIF-to-H\&E, (\textbf{h}) H\&E-to-mIF, and (\textbf{i}) Text-to-mIF. Haiku is compared against MUSK and a random-paired majority-voting baseline under identical evaluation settings.
    \textbf{j}, Schematic of CLIP-style zero-shot classification for mIF patches: organ-specific text prompts (illustrated here with ``A mIF image of breast/lung/colon/kidney'' as a subset of representative prompts) are encoded by the text encoder and matched against the mIF embedding of a query patch in the Haiku latent space, with the most similar prompt assigned as the predicted label.
    \textbf{k}, Zero-shot macro-averaged F1 for the mIF modality on organ-type (10 categories) and disease (11 categories) classification, covering the complete set of categories observed in the paired held-out test dataset (\textbf{Supplementary Figure~\ref{supplementary:3}}), comparing Haiku against a random-paired baseline (two-sided Wilcoxon rank-sum (Mann--Whitney $U$) test;  $^{*}P<0.05$, $^{**}P<0.01$, $^{***}P<0.001$).
    }
\end{figure}

\subsection{Haiku achieves state-of-the-art performance across diverse downstream clinical tasks at both patch and sample levels}

Haiku pretraining not only preserves unimodal features but also enhances representation by integrating multimodal information. To support these findings, we conduct patch-level linear probing evaluations against unimodal approaches (Methods~\ref{sec:linearprob}).

We first curate patch-level labels from clinical metadata in held-out tumor tissue slices for five cancer-related classification tasks: organ type, tissue type, and three tumor-related labels: T stage, N stage, and tumor grade (G1, G2, G3) (\textbf{Supplementary Figure~\ref{supplementary:3}}). We then perform tumor-related classification at the patch level for two reasons. First, our samples with cancer are derived from tissue microarrays (TMAs) rather than conventional whole-slide sections; therefore, most patches originate from tumor regions. Second, we assume that tumor stage and T/N status influence the tumor microenvironment, such that even non-tumor regions within the same slice may reflect stage-dependent differences.

We compare Haiku unimodal embeddings (Haiku(H\&E) and Haiku(mIF)) against fine-tuned VirTues and MUSK baselines representing unimodal state-of-the-art encoders, as well as a naive majority-voting baseline. We further hypothesize that integrating paired mIF and H\&E modalities can capture complementary information beyond a single modality. To test this hypothesis, we additionally concatenate Haiku(mIF) and Haiku(H\&E) embeddings to form a fused representation, denoted as Haiku(Fusion). All models are evaluated using five-fold cross-validation with linear probing (\textbf{Figure~\ref{fig:figure3}a}).

Across all tasks, Haiku unimodal embeddings consistently outperform unimodal baselines (MUSK for H\&E and VirTues for mIF) (\textbf{Figure~\ref{fig:figure3}b}). The Haiku(Fusion) embeddings further outperform all unimodal embeddings, achieving a macro F1 of 0.942 for N stage, 0.961 for T stage, 0.942 for tumor grade, 0.999 for organ types, and 0.998 for tissue types. All Haiku(Fusion) results are significantly better than the per-task second-best approach (two-sided paired $t$-test across the five folds; $P = 9.61\times10^{-5}$ for N stage, $P = 1.02\times10^{-4}$ for T stage, $P = 3.74\times10^{-6}$ for tumor grade, $P = 1.95\times10^{-2}$ for organ type, and $P = 6.98\times10^{-4}$ for tissue type), providing strong evidence that Haiku not only preserves unimodal features but also effectively integrates complementary information across modalities. 

Building on the evidence that Haiku captures patch-resolution tri-modal semantics in previous retrieval evaluation and generalizes to patch-level classification, we further investigate whether Haiku(mIF) embeddings capture clinically relevant sample-level information in more challenging downstream tasks (Methods~\ref{sec:mil}).

We evaluate Haiku on sample-level treatment response prediction and survival outcome prediction using the 198 unpaired mIF-only held-out slices introduced in the dataset section (\textbf{Figure~\ref{fig:figure1}b}), which comprise two held-out studies: a set of 75 metastatic melanoma samples from 75 unique patients (one acquisition per patient) with immunotherapy treatment information and detailed follow-up, and a set of 123 colorectal cancer (CRC) samples from 66 unique patients (multiple slices per patient) with treatment and longitudinal clinical outcome data. Both cohorts are entirely excluded from Haiku contrastive training and VirTues encoder pretraining, enabling a strict evaluation of real-world clinical generalization. Throughout, all five-fold cross-validation splits below are performed at the patient level rather than at the slice level, so that all slices from a given patient remain in the same fold and never simultaneously appear in both training and validation, thereby preventing patient-level leakage during evaluation.

For survival prediction, we train an attention-based multiple-instance learning (MIL) Cox regression model using five-fold cross-validation on the CRC cohort (\textbf{Figure~\ref{fig:figure3}c}). Each slice is treated as a bag, with instances corresponding to Haiku(mIF) or VirTues patch embeddings, and the bag-level label given by patient survival time. Haiku(mIF) embeddings achieve a higher mean concordance index (C-index) of $0.737$ compared with $0.683$ for the VirTues baseline, an improvement of approximately $0.054$ across the five folds (two-sided paired $t$-test; $P = 0.186$) (\textbf{Figure~\ref{fig:figure3}d}). Kaplan--Meier curves and log-rank tests further demonstrate clearer stratification between predicted low-risk and high-risk patient groups relative to baseline method VirTues. Specifically, VirTues yielded a log-rank $P$ value of $0.274$, whereas Haiku achieved a significantly stronger separation with a $P$ value of $3.41\times10^{-3}$ (\textbf{Figure~\ref{fig:figure3}e,f}).

We next evaluate treatment response prediction using MIL-based binary classification on both the melano-ma and CRC cohorts. Using the same MIL framework as in survival prediction and modifying only the training objective from Cox regression to binary classification, we compare Haiku(mIF) against VirTues across five-fold cross-validation, evaluated by both AUPRC (area under the precision--recall curve) and AUROC (area under the receiver operating characteristic curve). For the melanoma cohort, Haiku(mIF) achieves a mean AUROC of $0.756$ versus $0.352$ for VirTues and a mean AUPRC of $0.660$ versus $0.333$ for VirTues (two-sided paired $t$-test across the five folds; $P = 0.0097$ for AUROC and $P = 0.023$ for AUPRC) (\textbf{Figure~\ref{fig:figure3}g--i}). For the CRC cohort, Haiku(mIF) likewise achieves better mean values across five folds on both metrics, reaching a mean AUROC of $0.730$ versus $0.721$ for VirTues and a mean AUPRC of $0.775$ versus $0.735$ for VirTues (two-sided paired $t$-test; $P = 0.746$ for AUROC and $P = 0.166$ for AUPRC) (\textbf{Figure~\ref{fig:figure3}j--l}). Representative ROC and precision--recall curves from a single fold further illustrate the substantial improvement on melanoma (Haiku vs.\ VirTues single-fold AUROC $0.920$ vs.\ $0.320$; single-fold AUPRC $0.885$ vs.\ $0.308$; \textbf{Figure~\ref{fig:figure3}h,i}) and the more modest but consistent improvement on CRC (single-fold AUROC $0.799$ vs.\ $0.736$; single-fold AUPRC $0.880$ vs.\ $0.806$; \textbf{Figure~\ref{fig:figure3}k,l}). Haiku(mIF) consistently yields better mean AUROC and AUPRC than VirTues on both cohorts, with the improvement reaching statistical significance on melanoma, demonstrating that patch-level contrastive learning produces patch embeddings that transfer meaningfully to sample-level clinical prediction even on modestly sized external cohorts.

These results demonstrate that Haiku, trained using patch-level contrastive learning, can generate robust representations that reflect sample-level clinical properties and outcomes.

\begin{figure}[hbtp]
    \centering
    \includegraphics[width=\textwidth]{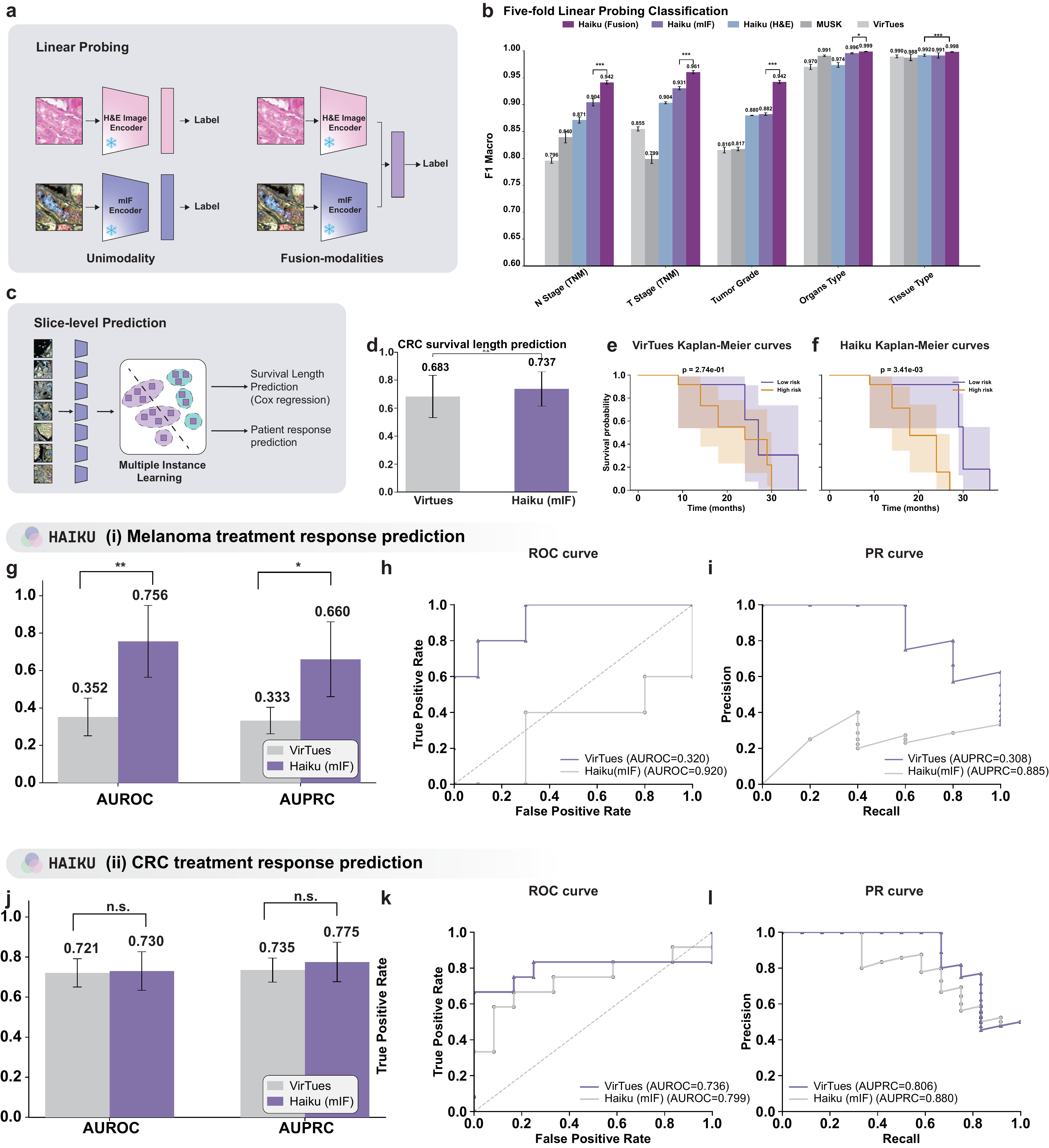}
    \captionsetup{
      format=plain,
      justification=raggedright,
      singlelinecheck=false,
      margin=0pt,
      width=\dimexpr\textwidth\relax
    }
    \caption[\textbf{Downstream linear probing and slice-level prediction tasks.}]%
    {\textbf{Downstream linear probing and slice-level prediction tasks.} See next page for caption.}
    \label{fig:figure3}
\end{figure}

\begin{figure}[hbtp]
    \ContinuedFloat
    \captionsetup{list=no} 
    \caption{
    \small
    \textbf{Downstream linear probing and slice-level prediction tasks.}
    \textbf{a}, Schematic of the linear-probing evaluation protocol. Frozen Haiku encoders (H\&E or mIF, marked with a snowflake) provide unimodal embeddings that are fed to a linear classifier under the unimodality setting, while fusion-modality evaluation concatenates Haiku(H\&E) and Haiku(mIF) embeddings before the classifier.
    \textbf{b}, Five-fold linear-probing classification on held-out tumor tissue slices across five clinically relevant tasks (N stage, T stage, tumor grade, organ type, and tissue type), reported as macro-averaged F1. Haiku(Fusion) is compared against Haiku(mIF), Haiku(H\&E), unimodal baselines (MUSK for H\&E, VirTues for mIF), and a majority-vote baseline (two-sided paired $t$-test between Haiku(Fusion) and the second-best method across the five folds; $^{*}P<0.05$, $^{**}P<0.01$, $^{***}P<0.001$).
    \textbf{c}, Schematic of the multiple-instance learning (MIL) protocol, in which pretrained Haiku(mIF) patch embeddings from each acquisition (TMA) are aggregated by an attention-based MIL head for both survival prediction (Cox regression) and treatment-response classification, with five-fold cross-validation performed at the patient level to prevent patient-level leakage.
    \textbf{d}, Five-fold mean concordance index (C-index) for colorectal cancer survival-length prediction, comparing VirTues and Haiku(mIF); Haiku achieves a better mean C-index (two-sided paired $t$-test; n.s.).
    \textbf{e--f}, Representative Kaplan--Meier curves stratified by median predicted risk (low vs.\ high) from a single fold with log-rank test for (\textbf{e}) VirTues and (\textbf{f}) Haiku, demonstrating clearer risk stratification by Haiku.
    \textbf{g--i}, Melanoma treatment-response prediction benchmarking: (\textbf{g}) AUROC and AUPRC summary bars with five-fold mean values comparing VirTues and Haiku(mIF); Haiku achieves better mean performance on both metrics (two-sided paired $t$-test;  $^{*}P<0.05$, $^{**}P<0.01$, $^{***}P<0.001$); (\textbf{h}) representative ROC curve from a single fold; (\textbf{i}) representative precision--recall curve from a single fold.
    \textbf{j--l}, Colorectal cancer (CRC) treatment-response prediction benchmarking: (\textbf{j}) AUROC and AUPRC summary bars with five-fold mean values comparing VirTues and Haiku(mIF); Haiku achieves better mean performance on both metrics (two-sided paired $t$-test;); (\textbf{k}) representative ROC curve from a single fold; (\textbf{l}) representative precision--recall curve from a single fold.
    }
\end{figure}

\subsection{Zero-shot fusion retrieval enables metadata-conditioned biomarker inference}

Having demonstrated that Haiku captures semantic alignment between image modalities, we next investigate whether incorporating clinical metadata can improve biomarker inference beyond what H\&E embeddings achieve alone. A fundamental limitation of unimodal retrieval is that each modality carries incomplete information: H\&E images capture tissue morphology, whereas clinical text encodes disease, staging, and outcome information. To exploit this complementarity, we introduce \emph{fusion retrieval} (Methods~\ref{sec:biomarker}), which combines the H\&E and text similarity scores against the mIF reference atlas through a weighted sum before ranking (\textbf{Figure~\ref{fig:figure4}a}). To isolate the contribution of clinical context, we construct \emph{metadata-only} text descriptions that retain tissue-level clinical information (organ type, disease status, staging) but deliberately exclude all explicit biomarker abundance information (Methods~\ref{sec:biomarker}), so that any improvement over H\&E-only retrieval arises from complementary semantic knowledge rather than direct molecular supervision.

We evaluate biomarker inference accuracy using the Pearson correlation coefficient (PCC) between the similarity-weighted predicted and ground-truth mIF biomarker abundance profiles, computed independently for each of 52 validated biomarker channels across 336 held-out tissue regions (see Methods~\ref{sec:biomarker} for the full aggregation procedure). We compare four retrieval strategies: Haiku(H\&E), which uses H\&E embeddings alone; Haiku(Text), which uses metadata-only text embeddings; Haiku(Fusion), which combines H\&E and metadata-only text embeddings with fusion weights ($\alpha = 0.8$ for H\&E, $1-\alpha = 0.2$ for text); and the MUSK baseline.

Haiku(Fusion) achieves the highest mean PCC of $0.718$, significantly outperforming Haiku(H\&E) alone ($0.710$; two-sided Wilcoxon signed-rank test, $P = 1.46\times10^{-5}$) (\textbf{Figure~\ref{fig:figure4}b}). Despite the absence of any biomarker information in the text query, the fusion strategy consistently improves retrieval-based biomarker prediction, indicating that clinical metadata encoded through tri-modal alignment provides complementary information beyond morphology. The same fusion mechanism also improves direct cross-modality retrieval itself, raising Recall@$50$ from $0.611$ under H\&E-only queries to $0.643$ under fused H\&E$+$text queries while text-only queries reach only $0.169$ (\textbf{Supplementary Figure~\ref{supplementary:4}}). Both Haiku strategies vastly outperform MUSK (mean PCC $-0.033$), which produces near-zero or negative correlations; this large gap reflects the fact that MUSK is a vision encoder pretrained on three-channel RGB H\&E images, so accommodating it for mIF requires collapsing the many biomarker channels of each CODEX acquisition into an RGB-like three-channel input, which discards most of the per-channel molecular signal that biomarker inference relies on. This comparison highlights the necessity of a dedicated mIF encoder paired with explicit cross-modal alignment, rather than re-purposing a single-modality H\&E backbone, for retrieval-based biomarker inference from multiplexed protein imaging.

Per-biomarker analysis reveals robust improvement across diverse biological programs (\textbf{Figure~\ref{fig:figure4}c--g}; see \textbf{Supplementary Figure~\ref{supplementary:5}} for exact per-biomarker PCC values for all four methods), spanning adaptive immune markers (CD3e, CD4, CD8, CD20, PD1, PDL1), tumor-intrinsic markers (EpCAM, PanCK, PCNA, Ki67), and stromal components (Collagen~IV, Podoplanin, CD31). This breadth indicates that tri-modal alignment transfers fine-grained molecular information across functional categories, substantially expanding the scope of biological insights inferable from H\&E images augmented with clinical context. Qualitative retrieval examples for four representative biomarkers (CD11b, ICOS, PDL1, and PGP9\_5) confirm that retrieved mIF patches closely match ground-truth spatial distributions (\textbf{Figure~\ref{fig:figure4}h--k}).

\begin{figure}[hbtp]
    \centering
    \includegraphics[width = \textwidth]{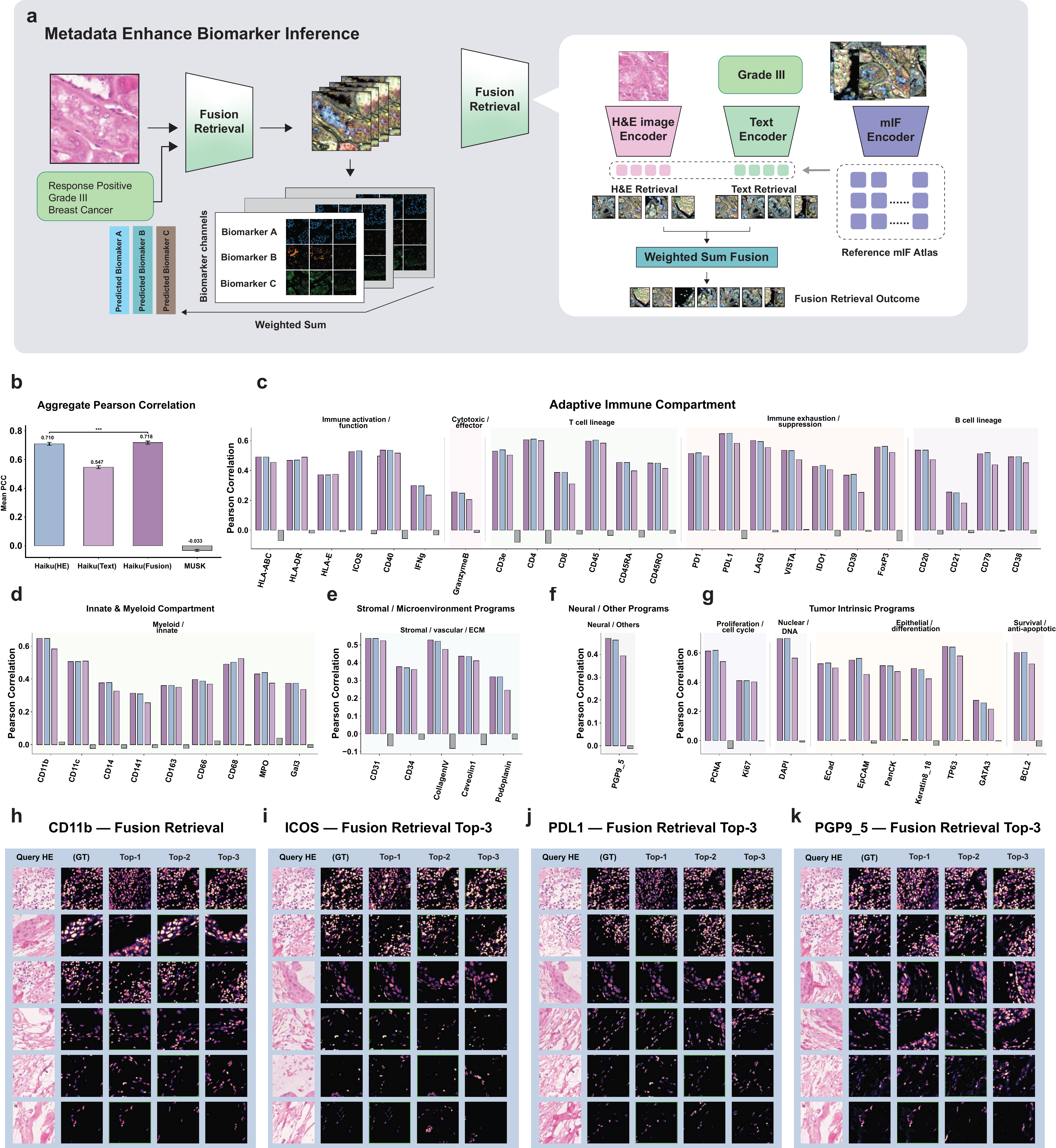}
    \captionsetup{
      format=plain,
      justification=raggedright,
      singlelinecheck=false,
      margin=0pt,
      width=\dimexpr\textwidth\relax
    }
    \caption[\textbf{Zero-shot fusion retrieval--based biomarker inference.}]%
    {\textbf{Zero-shot fusion retrieval--based biomarker inference.} See next page for caption.}
    \label{fig:figure4}
\end{figure}

\begin{figure}[hbtp]
    \ContinuedFloat
    \captionsetup{list=no}
    \caption{%
    \small
    \textbf{Zero-shot fusion retrieval--based biomarker inference.}
    \textbf{a}, Schematic of metadata-enhanced biomarker inference via fusion retrieval. A query H\&E patch and its paired metadata-only text (retaining clinical context but excluding biomarker information) are encoded separately; their similarities to a reference mIF atlas are linearly combined with optimized weights to produce a fusion retrieval result, from which predicted biomarker abundances are computed as a similarity-weighted sum over retrieved mIF patches.
    \textbf{b}, Aggregate mean Pearson correlation coefficient (PCC) between predicted and ground-truth biomarker abundance, comparing four retrieval strategies: Haiku(H\&E), Haiku(Text) using metadata-only descriptions, Haiku(Fusion) combining H\&E and metadata-only text with optimized weights ($0.8$ H\&E $+$ $0.2$ Text), and the MUSK baseline. Results are aggregated across 52 biomarkers and 336 held-out regions (two-sided paired Wilcoxon signed-rank test; $^{*}P<0.05$, $^{**}P<0.01$, $^{***}P<0.001$).
    \textbf{c--g}, Per-biomarker PCC grouped by biological program, comparing Haiku(H\&E), Haiku(Text), Haiku(Fusion), and MUSK: (\textbf{c}) adaptive immune compartment (immune activation/function, cytotoxic/effector, T-cell lineage, immune exhaustion/suppression, and B-cell lineage markers); (\textbf{d}) innate and myeloid compartment; (\textbf{e}) stromal and microenvironment programs (stromal, vascular, and extracellular matrix (ECM) markers); (\textbf{f}) neural and other programs; and (\textbf{g}) tumor-intrinsic programs (proliferation and cell cycle, nuclear/DNA, epithelial/differentiation, and survival/anti-apoptotic markers). The breadth of improvement across these functional categories demonstrates that tri-modal alignment transfers fine-grained molecular information spanning diverse biological programs.
    \textbf{h--k}, Qualitative fusion retrieval examples for four representative biomarkers---(\textbf{h}) CD11b, (\textbf{i}) ICOS, (\textbf{j}) PDL1, and (\textbf{k}) PGP9\_5---showing, for each query H\&E patch, the ground-truth mIF biomarker channel and the corresponding top-3 retrieved mIF patches, illustrating the spatial fidelity of fusion-retrieval-based biomarker inference.
    }
\end{figure}

\subsection{Metadata-only counterfactual prediction reveals niche-specific cancer progression dynamics}

A capability afforded by Haiku's tri-modal alignment of H\&E, mIF, and clinical text in a shared embedding space is zero-shot \emph{counterfactual perturbation}. Prior computational pathology studies have largely been confined to unidirectional mappings, such as H\&E predicts disease/outcome label~\cite{Chen2024-sl,Wang2024-hs,Xu2024-do,Ding2025-mv}, mIF predicts disease/outcome label~\cite{Wenckstern2025-ve,shaban2025foundation,liu2025modeling}, or H\&E predicts mIF~\cite{Wu2025-nj,Valanarasu2025-ek,Li2026-kw}, which permit inference within a fixed input modality but cannot interrogate how a controlled change in one modality propagates to the others. Because our model embeds all three modalities into a unified space, we can instead hold one modality fixed, perturb another, and compare the resulting retrieval against the unperturbed baseline to see how the third modality shifts.

Specifically, given a real H\&E patch paired with its own clinical metadata, we first perform fusion retrieval using the H\&E patch and its original metadata text to obtain an \emph{original} mIF retrieval set (\textbf{Figure~\ref{fig:figure5}a}). We then alter a single metadata attribute while keeping the H\&E morphology fixed, and re-query the atlas to obtain a \emph{counterfactual} mIF set (\textbf{Figure~\ref{fig:figure5}a}; Methods~\ref{sec:counterfactual_analysis}). For example, we can perturb ``tumor stage'' from T2 to T4 and observe how the retrieved mIF set changes. Comparing the retrieved biomarker-expression patterns between original and counterfactual sets allows us to examine differences in the molecular profiles of TMEs under different clinical conditions, which may surface hypotheses for follow-up investigation.

Haiku, with its jointly trained tri-modal embedding space, makes this perturb-and-retrieve paradigm practical to apply: text edits map onto retrieval shifts in the shared latent space, and the diversity of the reference atlas helps ensure that retrieved counterfactual neighbors reflect biological variation present in the data rather than nearest-available artifacts. Unlike unimodal retrieval, which cannot jointly ingest morphology and a text prompt, the tri-modal embedding anchors each query within tissue context while allowing single-attribute perturbations of the clinical metadata. We use this setup as a hypothesis-generating tool that complements existing generative or predictive modeling approaches, with all results below interpreted as exploratory and requiring further validation.

\begin{figure}[!ht]
    \centering
    \includegraphics[width=0.95\textwidth]{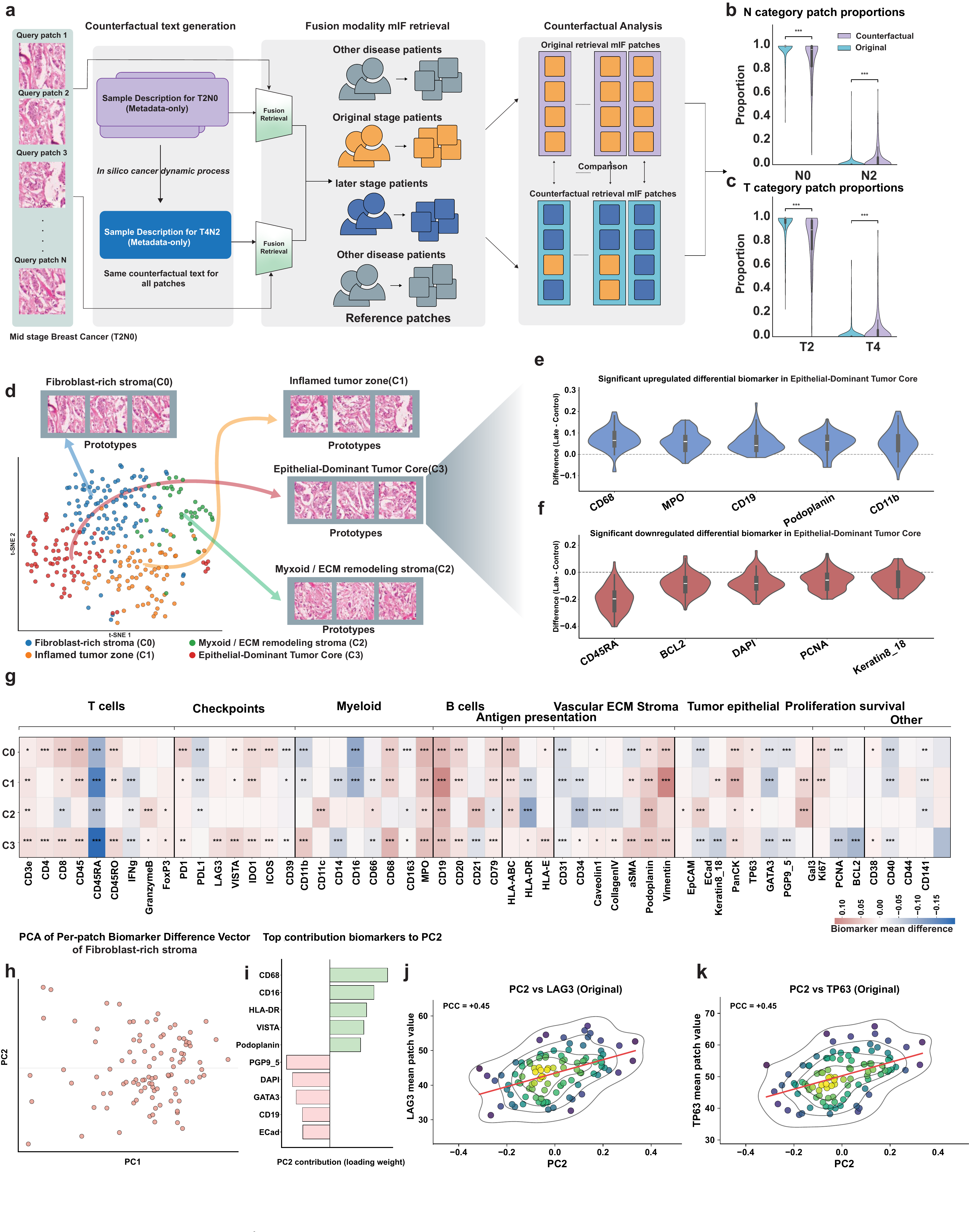}
        \captionsetup{
          format=plain,
          justification=raggedright,
          singlelinecheck=false,
          margin=0pt,
          width=\dimexpr\textwidth\relax
        }
    \caption[\textbf{Metadata-only counterfactual analysis of breast cancer progression dynamics.}]%
    {\textbf{Metadata-only counterfactual analysis of breast cancer progression dynamics.} See next page for caption.}
    \label{fig:figure5}
\end{figure}

\begin{figure}[!t]
\ContinuedFloat
\captionsetup{list=no}

\caption{%
\small
\textbf{Metadata-only counterfactual analysis of breast cancer progression dynamics.}
\textbf{a}, Overview of the zero-shot counterfactual prediction workflow simulating \emph{in silico} progression from mid-stage ($T2N0$) to late-stage ($T4N2$) breast cancer. For each H\&E query patch, fusion retrieval is first performed using the original metadata-only text description (containing clinical context but excluding biomarker information) against a reference atlas composed of original-stage patients, later-stage patients, and other-disease patients to produce the original mIF retrieval set; replacing only the staging fields in the metadata text with a counterfactual late-stage description, while keeping the H\&E embedding and all other clinical fields fixed, generates the counterfactual retrieval set under otherwise identical constraints, enabling direct comparison between the two sets.
\textbf{b--c}, Metadata population shifts from the single T2N0$\to$T4N2 perturbation, shown as two complementary views of the same retrieved mIF patch sets: (\textbf{b}) within-query proportions of retrieved patches assigned to each $N$-stage category (N0, N2) and (\textbf{c}) to each $T$-stage category (T2, T4), displayed as violin plots over the paired $n=281$ queries (paired two-sided Wilcoxon rank-sum (Mann--Whitney $U$) test with Benjamini--Hochberg false discovery rate (FDR) control across categories; adjusted $^{*}P<0.05$, $^{**}P<0.01$, $^{***}P<0.001$; Methods~\ref{sec:subpop_shift}).
\textbf{d}, t-SNE visualization of Haiku(H\&E) embeddings for the control breast cancer query patches, colored by $K$-means clusters ($K=4$; Methods~\ref{sec:he_microenv_cluster}), with representative H\&E prototypes annotated for each of the four morphological compartments: fibroblast-rich stroma (C0), inflamed tumor zone (C1), myxoid/ECM remodeling stroma (C2), and epithelial-dominant tumor core (C3).
\textbf{e--f}, Violin plots of the most significantly (\textbf{e}) upregulated and (\textbf{f}) downregulated differential biomarkers (counterfactual $-$ original weighted mean abundance) within the epithelial-dominant tumor core (C3), where ``significantly upregulated'' biomarkers have mean per-query shifts significantly greater than zero and ``significantly downregulated'' biomarkers have mean per-query shifts significantly less than zero (two-sided Wilcoxon signed-rank test against zero with Benjamini--Hochberg FDR control across biomarkers; adjusted $P<0.05$; Methods~\ref{sec:cluster_biomarker_test}). 
\textbf{g}, Heatmap of group-wise mean biomarker abundance differences (counterfactual $-$ original) across all four morphological compartments (C0--C3), organized by biological program (T cells, checkpoints, myeloid, B cells, antigen presentation, vascular/ECM/stroma, tumor/epithelial, proliferation/survival, and other); color encodes the mean biomarker abundance difference (counterfactual $-$ original) per cluster, and significance stars reflect the same two-sided Wilcoxon signed-rank test against zero with Benjamini--Hochberg FDR control used for panels e and f (adjusted $^{*}P<0.05$, $^{**}P<0.01$, $^{***}P<0.001$; Methods~\ref{sec:cluster_biomarker_test}).
\textbf{h}, Principal component analysis (PCA) of per-patch biomarker difference vectors in the fibroblast-rich stroma (C0), with each point representing one query patch in the PC1--PC2 space.
\textbf{i}, Top-weighted biomarkers contributing to PC2, with the sign of the loading indicating direction of association.
\textbf{j--k}, Correlation between PC2 scores and baseline ground-truth biomarker values in the control matched mIF patches for (\textbf{j}) the immune-checkpoint marker LAG3 (PCC $= +0.45$) and (\textbf{k}) the basal/myoepithelial marker TP63 (PCC $= +0.45$), indicating that within-cluster baseline immune-checkpoint and basal-lineage state are positively associated with the leading PC2 axis of counterfactual change. Scatter point color encodes the local two-dimensional kernel density estimate of the joint distribution (PC2, biomarker), with warmer colors indicating higher local point density.
}
\end{figure}

As a first counterfactual, we investigate cancer progression dynamics. We performed counterfactual retrieval on breast cancer patches drawn from the 336-slice paired held-out dataset (\textbf{Figure~\ref{fig:figure1}b}). Specifically, we used all 281 H\&E query patches tiled from a single mid-stage (T2N0M0, stage IIA, grade 2) breast cancer patient, and modified only the staging fields to (T4N2M1, stage IV, grade 3) in the metadata-only text descriptions, keeping all other clinical information and H\&E embeddings intact (\textbf{Figure~\ref{fig:figure5}a}); restricting the analysis to a single patient eliminates inter-patient morphological heterogeneity as a confounder and isolates the molecular shifts attributable to the staging intervention. \textbf{Figure~\ref{fig:figure5}b,c} summarize the outcome of this single T2N0$\to$T4N2 perturbation from two complementary views onto the same retrieved mIF patch sets: the $N$-stage label composition (\textbf{Figure~\ref{fig:figure5}b}) and the $T$-stage label composition (\textbf{Figure~\ref{fig:figure5}c}). Along the $N$-stage axis, the within-query proportion of retrieved patches labeled $N0$ decreased from a mean of $0.966$ under the original metadata to $0.886$ under the counterfactual metadata, while the proportion labeled $N2$ increased from $0.013$ to $0.049$ (paired $n=281$ queries; paired two-sided Wilcoxon rank-sum  test with Benjamini--Hochberg FDR control; adjusted $P = 6.8\times10^{-27}$ for $N0$ and $P = 4.3\times10^{-22}$ for $N2$; Methods~\ref{sec:subpop_shift}). Along the $T$-stage axis the shifts are similarly significant: the proportion of $T2$-labeled retrievals drops from $0.946$ to $0.827$ and the proportion of $T4$-labeled retrievals rises from $0.019$ to $0.061$ (same paired $n=281$; adjusted $P = 2.8\times10^{-33}$ for $T2$ and $P = 1.5\times10^{-17}$ for $T4$; Methods~\ref{sec:subpop_shift}). These results indicate that the fusion retrieval is responsive to the staging prompt rather than returning invariant results.

Next, we assessed how counterfactual perturbation affected marker expression in tissues. To localize such changes within morphologically distinct niches, we first stratified the 281 query patches into four compartments by $K$-means clustering ($K=4$) of their Haiku(H\&E) embeddings and annotated each cluster based on representative H\&E patch appearance (see Methods~\ref{sec:he_microenv_cluster} for details): fibroblast-rich stroma (C0, $n=100$ patches), inflamed tumor zone (C1, $n=70$), myeloid/ECM remodeling stroma (C2, $n=39$), and epithelial-dominant tumor core (C3, $n=72$) (\textbf{Figure~\ref{fig:figure5}d}). Within each cluster, we then compared the counterfactual-minus-original weighted mean biomarker abundance per query patch using a two-sided Wilcoxon signed-rank test against zero, with Benjamini--Hochberg FDR control across the 52 biomarkers (Methods~\ref{sec:cluster_biomarker_test}). Per-marker shifts are reported in the form (\textit{percentage change}, $P$-value), where the percentage change is relative to the original per-patch mean abundance and $P$ is the FDR-adjusted significance bin ($P<0.001$, $P<0.01$, or $P<0.05$). In the \emph{epithelial-dominant tumor core} (C3), the counterfactual late-stage perturbation produces two coordinated, niche-appropriate shifts (\textbf{Figure~\ref{fig:figure5}e--f}): increases in the pan-macrophage marker CD68 ($+69.7\%$, $P<0.001$) and the lymphatic/cancer-associated-fibroblast marker Podoplanin ($+99.9\%$, $P<0.001$), both directly reported in breast cancer to be associated with advanced histological grade and poor prognosis~\cite{Ni2019Meta,Schoppmann2012Podoplanin}, alongside coordinated losses of the canonical luminal-defining trio GATA3 ($-22.3\%$), Keratin8\_18 ($-23.0\%$), and E-cadherin ($-13.4\%$) (all $P<0.001$), which directly mirror the loss of luminal differentiation that accompanies breast cancer progression and dissemination~\cite{Kouros-Mehr2008GATA3,AschKendrick2016GATA3}.

Examining the per-compartment biomarker shifts across the full heatmap (\textbf{Figure~\ref{fig:figure5}g}; same Wilcoxon test with FDR control; Methods~\ref{sec:cluster_biomarker_test}) reveals that the counterfactual response takes distinct forms in each microenvironment compartment. In the \emph{fibroblast-rich stroma} (C0), the dominant signal is broad adaptive-immune infiltration, with coordinated gains in B-cell markers (CD19 $+70.5\%$, CD20 $+132.9\%$, CD79 $+142.7\%$; all $P<0.001$) and a CD8 T-cell signal (CD8 $+28.9\%$, $P<0.001$), accompanied by losses of vasculature and smooth-muscle stromal structure. In the \emph{inflamed tumor zone} (C1), the perturbation produces the strongest mesenchymal signal in the panel: Vimentin is the top-ranked upregulated marker ($+73.9\%$, $P<0.001$), co-occurring with significant downregulation of the master luminal transcription factor GATA3 ($-41.1\%$, $P<0.001$). The direction of this Vimentin$\uparrow$/GATA3$\downarrow$ shift is consistent with the experimental finding that GATA3 loss in breast cancer cells drives the canonical epithelial-to-mesenchymal transition (EMT)~\cite{Yan2010GATA3EMT,Kouros-Mehr2008GATA3}. In the \emph{myxoid/ECM remodeling stroma} (C2), the dominant signal is loss of antigen presentation: HLA-DR is the top-ranked downregulated marker ($-34.2\%$, $P<0.001$), accompanied by concurrent vasculature and basement-membrane collagen losses; loss of intratumoral HLA-DR has been directly reported to associate with poorer outcomes in triple-negative breast cancer~\cite{Wang2023HLADR}, so the recovered direction in C2 is consistent with a worse-prognosis-associated antigen-presentation collapse alongside compartment-specific stromal-niche reorganization. The four compartments show qualitatively distinct counterfactual programs from the same metadata perturbation.

Within this cross-compartment heatmap (\textbf{Figure~\ref{fig:figure5}g}), the magnitudes of individual marker shifts also vary substantially by compartment. As one representative example, the naive T-cell marker CD45RA is significantly downregulated in all four clusters (C0 $-29.5\%$, C1 $-49.7\%$, C2 $-39.8\%$, C3 $-43.9\%$; all $P<0.001$), with the largest depletion in the morphologically tumor-bearing niches (C3 and C1) and the smallest in the dedicated stromal compartment (C0) (\textbf{Figure~\ref{fig:figure5}g}, ``T cells'' row). This direction is consistent with previous breast cancer studies reporting that naive CD45RA$^+$ T cells are depleted in tumor-bearing tissue~\cite{Egelston2018BC}, and with the broader studies that the tumor immune compartment shifts away from naive/quiescent populations as disease progresses~\cite{Goff2015BCprogression,Hu2017CD45RO}. The cross-compartment magnitude gradient further confirms that Haiku resolves compartment-specific responses to the same metadata perturbation.

To examine within-compartment heterogeneity of the counterfactual response, we performed principal component analysis on the per-patch biomarker difference vectors within the \emph{fibroblast-rich stroma} (C0; \textbf{Figure~\ref{fig:figure5}h}), the largest compartment. PC2 separates patches whose counterfactual response is contributed positively by myeloid and antigen-presentation markers (top positive loadings: CD68, CD16, HLA-DR, VISTA, Podoplanin) from those contributed negatively by epithelial and B-lineage markers (top negative loadings: ECad, CD19, GATA3, DAPI, PGP9\_5; \textbf{Figure~\ref{fig:figure5}i}). When PC2 scores are correlated against pre-perturbation biomarker abundance, two markers emerge with the strongest positive correlation: the immune-checkpoint marker LAG3 and the basal/myoepithelial marker TP63 (both PCC $= +0.45$; \textbf{Figure~\ref{fig:figure5}j,k}). This indicates that the direction of an individual patch's counterfactual response within C0 is conditioned on its baseline immune-checkpoint and basal-lineage state, with patches of higher baseline LAG3 and TP63 trending toward the myeloid-leaning end of the response axis. Together, these results show that Haiku's counterfactual framework reveals both compartment-specific and patch-conditioned biomarker shifts from a metadata-only perturbation.

\subsection{Counterfactual prediction of survival-associated microenvironment remodeling in lung cancer}
 
\begin{figure}[hbtp]
    \centering
    \includegraphics[width=\textwidth]{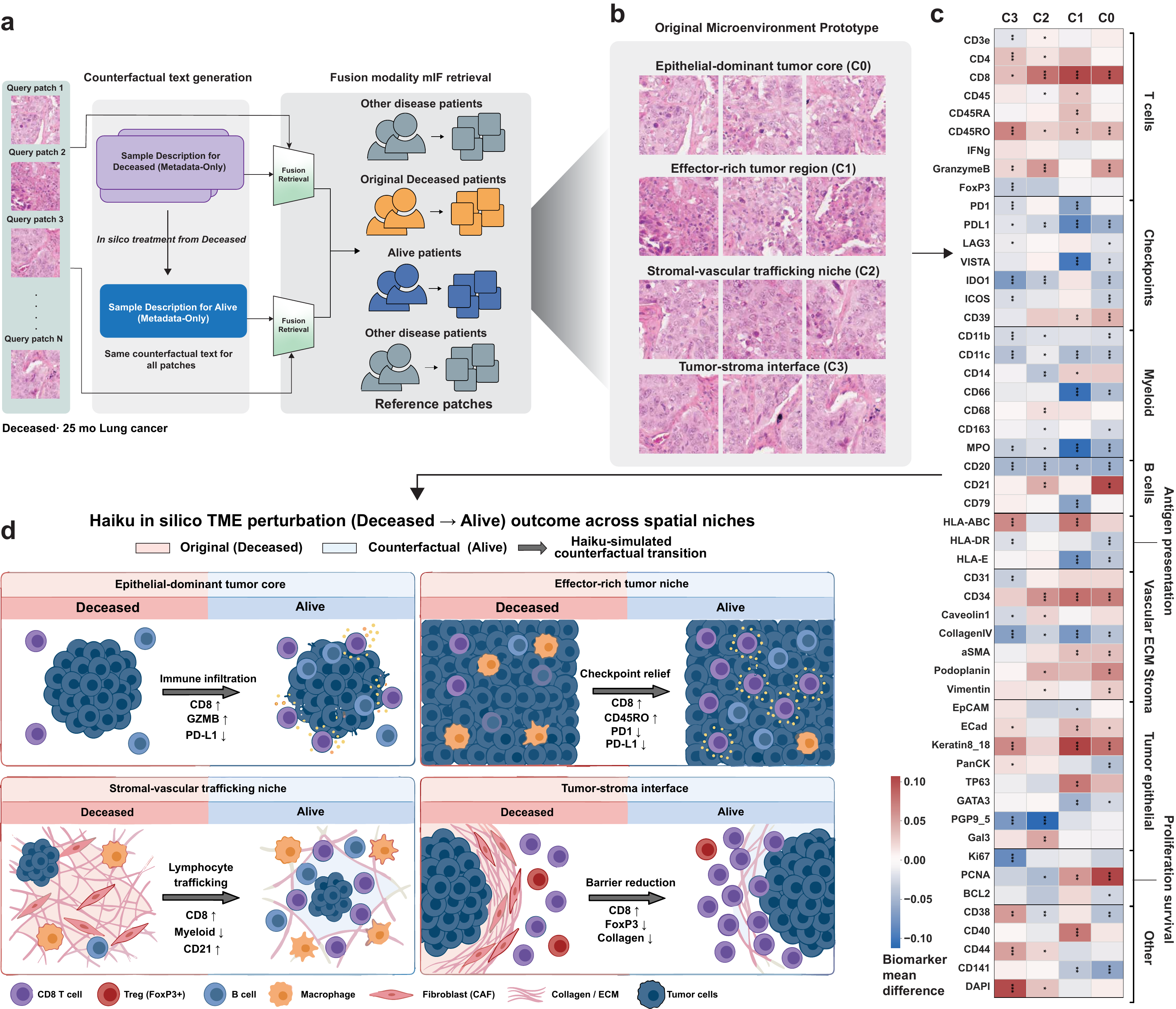}
    \caption{
    \small
    \textbf{Metadata-only counterfactual analysis of survival outcome in lung cancer.}
    \textbf{a}, Overview of the zero-shot counterfactual workflow for an \emph{in silico} survival intervention on a single lung adenocarcinoma patient (Deceased, survival 25 months). For each H\&E query patch, fusion retrieval is performed using the original metadata-only text (``Deceased'') against a reference atlas containing original deceased patients, alive patients, and other-disease patients, producing the original mIF retrieval set; replacing only the survival status field with ``Alive'' in the metadata text, while keeping the H\&E embedding and all other clinical fields fixed, produces the counterfactual retrieval set for comparison.
    \textbf{b}, H\&E-derived microenvironment prototypes from $K$-means clustering of Haiku(H\&E) embeddings, identifying four spatial niches with three representative H\&E patches each: epithelial-dominant tumor core (C0), effector-rich tumor niche (C1), stromal-vascular trafficking niche (C2), and tumor--stroma interface (C3).
    \textbf{c}, Heatmap of biomarker abundance differences (Alive $-$ Deceased) for all four spatial niches (columns C3, C2, C1, C0), organized along the rows by biological program: T cells, checkpoints, myeloid, B cells, antigen presentation, vascular/ECM/stroma, tumor/epithelial, proliferation/survival, and other (two-sided Wilcoxon signed-rank test against zero with Benjamini--Hochberg FDR control across biomarkers; adjusted $^{*}P<0.05$, $^{**}P<0.01$, $^{***}P<0.001$; color indicates the mean biomarker difference per cluster; Methods~\ref{sec:cluster_biomarker_test}).
    \textbf{d}, Schematic summary of niche-specific remodeling programs inferred from the counterfactual prediction. Each panel contrasts the original (Deceased) and counterfactual (Alive) cell-population composition together with the dominant biomarker shifts and their biological interpretation: immune infiltration in the epithelial-dominant tumor core (C0; CD8$\uparrow$, GZMB$\uparrow$, PD-L1$\downarrow$), checkpoint relief in the effector-rich tumor niche (C1; CD8$\uparrow$, CD45RO$\uparrow$, PD1$\downarrow$, PD-L1$\downarrow$), lymphocyte trafficking in the stromal-vascular niche (C2; CD8$\uparrow$, CD21$\uparrow$, myeloid$\downarrow$), and barrier reduction at the tumor--stroma interface (C3; CD8$\uparrow$, FoxP3$\downarrow$, collagen$\downarrow$). All arrows correspond to biomarker shifts that reach FDR significance in the cluster-stratified test (see panel c and Methods~\ref{sec:cluster_biomarker_test}). Cell-type legend: CD8 T cell, Treg (FoxP3$^+$), B cell, macrophage, fibroblast (CAF), collagen/ECM, and tumor cells.
    }
    \label{fig:figure6}
\end{figure}

Next, we explored whether Haiku's counterfactual framework can be applied to clinically actionable questions such as inferring molecular determinants of patient survival. As an example, we used all 154 H\&E query patches tiled from a single lung adenocarcinoma patient with deceased outcome (survival: 25 months, stage IIIA, T3N1M0) drawn from the 336-slice paired held-out dataset (\textbf{Figure~\ref{fig:figure1}b}), and modified only the survival status from ``Deceased'' to ``Alive'' in the metadata-only text description while keeping all other clinical fields and the H\&E embeddings fixed (\textbf{Figure~\ref{fig:figure6}a}). As in the breast cancer analysis, restricting the perturbation to all patches from one patient fixes the underlying morphology and clinical context so that the counterfactual shifts can be attributed to the survival-status intervention rather than to cross-patient variability. This \emph{in silico} survival perturbation aims to gauge whether and what molecular microenvironment remodeling would distinguish favorable from unfavorable outcomes under identical morphological and clinical contexts. 
 
We stratified the 154 patches into four spatial niches by $K$-means clustering ($K=4$) of their Haiku(H\&E) embeddings, with each cluster manually annotated based on representative H\&E prototypes (Methods~\ref{sec:he_microenv_cluster}; \textbf{Figure~\ref{fig:figure6}b}): \emph{epithelial-dominant tumor core} (C0, $n=42$ patches), \emph{effector-rich tumor niche} (C1, $n=30$), \emph{stromal-vascular trafficking niche} (C2, $n=31$), and \emph{tumor--stroma interface} (C3, $n=51$). Within each niche, we compared biomarker abundance between the original (Deceased) and counterfactual (Alive) retrieval sets using the same Wilcoxon test with FDR control (Methods~\ref{sec:cluster_biomarker_test}); per-marker shifts are reported in the form (\textit{percentage change}, $P$-value), where the percentage change is relative to the ground-truth (original) per-patch mean abundance and $P$ is the FDR-adjusted significance bin ($P<0.001$, $P<0.01$, or $P<0.05$). The complete biomarker-by-niche shift is shown in \textbf{Figure~\ref{fig:figure6}c}, with the dominant niche-specific shifts summarized schematically in \textbf{Figure~\ref{fig:figure6}d} and detailed below.
 
In the \emph{epithelial-dominant tumor core} (C0), the counterfactual Alive state shows coordinated immune infiltration (\textbf{Figure~\ref{fig:figure6}c}, column C0), with significantly increased cytotoxic markers (CD8 $+50.6\%$, $P<0.001$; granzyme B $+38.0\%$, $P<0.001$) and the memory T-cell marker CD45RO ($+36.8\%$, $P<0.001$) alongside significantly reduced immune checkpoint expression (PD-L1 $-61.7\%$, $P<0.001$), consistent with the established association between intratumoral CD8$^+$ T-cell density, memory T-cell enrichment, and favorable outcomes in non-small cell lung cancer (NSCLC)~\cite{Tumeh2014PDL1,Galon2006ImmuneTumor,Fridman2012ImmuneContexture,Pages2005CD45RO}. The \emph{effector-rich tumor niche} (C1) exhibits the most pronounced checkpoint relief program (\textbf{Figure~\ref{fig:figure6}c}, column C1), with significant reductions across co-inhibitory receptors PD1 ($-24.6\%$, $P<0.001$), PDL1 ($-30.7\%$, $P<0.001$), and VISTA ($-34.7\%$, $P<0.001$), co-occurring with expansion of effector and memory T-cell populations (CD8 $+89.5\%$, $P<0.001$; CD45RO $+35.7\%$, $P<0.01$) and clearance of the suppressive myeloid markers CD11c ($-35.6\%$, $P<0.001$) and MPO ($-27.5\%$, $P<0.001$); this multi-checkpoint shift is reminiscent of response signatures reported in immune checkpoint blockade therapy~\cite{Pardoll2012Checkpoint,Anderson2016LAG3,Herbst2018Atezolizumab}. Concurrent decreases in the epithelial/luminal markers EpCAM ($-16.1\%$, $P<0.05$) and GATA3 ($-27.9\%$, $P<0.01$) in this niche further suggest reduced tumor-cell presence accompanying the immune activation.
 
The \emph{stromal-vascular trafficking niche} (C2) shows lymphocyte recruitment signatures (\textbf{Figure~\ref{fig:figure6}c}, column C2) with significantly increased CD8 ($+35.8\%$, $P<0.001$) and the follicular B-cell marker CD21 ($+71.9\%$, $P<0.01$) together with significantly reduced granulocytic myeloid activity (MPO $-32.1\%$, $P<0.05$), consistent with a gateway function for effector cell recruitment into the tumor bed~\cite{Salmon2012MatrixArchitecture,Mantovani2017TAM}. Notably, the pan-B-cell marker CD20 decreases significantly in this niche ($-59.7\%$, $P<0.001$), so the B-cell shift is best interpreted as enrichment of a CD21$^+$ follicular/germinal-center-like subset rather than broad B-cell expansion, reminiscent of intratumoral tertiary lymphoid structures associated with favorable outcomes in NSCLC~\cite{Dieu-Nosjean2008TLS,Cabrita2020TLS}. Finally, the \emph{tumor--stroma interface} (C3) exhibits barrier reduction (\textbf{Figure~\ref{fig:figure6}c}, column C3) with significantly increased CD8 infiltration ($+13.5\%$, $P<0.05$), decreased regulatory T-cell suppression (FoxP3 $-36.1\%$, $P<0.001$), decreased extracellular matrix deposition (Collagen~IV $-23.7\%$, $P<0.001$), and the only FDR-significant reduction in tumor proliferation across the four niches (Ki67 $-24.8\%$, $P<0.001$; Ki67 trends same-sign but non-significant in C0, C1, and C2), a pattern that is broadly consistent with a shift away from an immunosuppressive, fibrotic boundary toward a more permissive interface for immune engagement~\cite{Joyce2015TMEExclusion,Kalluri2006Fibrosis}.
 
The per-niche programs recovered here align broadly with the schematic summary in \textbf{Figure~\ref{fig:figure6}d}, which contrasts the Deceased and Alive cell-population composition in each niche and captures the dominant biomarker shift in each: immune infiltration in C0, checkpoint relief in C1, lymphocyte trafficking in C2, and barrier reduction at the tumor--stroma interface in C3. At a high level, the counterfactual Alive state surfaces a coherent program with four convergent themes---effector T-cell expansion, broad checkpoint relief, suppressive myeloid clearance, and a niche-specific reduction in tumor proliferation that reaches FDR significance only at the \emph{tumor--stroma interface} (C3), with different niches emphasizing distinct aspects of this shared program. The directional consistency of these shifts, derived solely from a metadata-only survival-status intervention, is suggestive of a coordinated immune-activation axis associated with favorable prognosis. We emphasize that this analysis is a single-patient proof-of-concept demonstration; extending the framework to larger cohorts with systematic cross-patient validation will be necessary to assess the generalizability of any recovered program. With those limitations in mind, these results illustrate one way in which Haiku's counterfactual framework might be used as an exploratory, hypothesis-generation tool for surfacing candidate survival-associated molecular signals for follow-up investigation.

\section{Discussion}
\label{sec:discussion}
Although unimodal foundation models have recently achieved strong performance on H\&E images~\cite{Chen2024-sl, Xu2024-do} and on spatial omics modalities such as mIF~\cite{Wenckstern2025-ve, shaban2025foundation, liu2025modeling} and spatial transcriptomics~\cite{wang2025scgptspatial, schaar2025nicheformer}, and several integrative efforts have begun to align histology with spatial omics~\cite{Chen2025-ee, Huang2025-bv}, there remains no model that connects the flexible, human-interpretable text modality with these complex, structured spatial omics and imaging modalities within a single shared representation.

Our work presents \textbf{Haiku}, the first tri-modal foundation model that leverages one of the largest spatial proteomics atlases and jointly aligns spatial proteomics, H\&E histology images, and textual metadata information within a shared embedding space.

We perform extensive large-scale evaluations on held-out datasets to assess cross-modality alignment and demonstrate, for the first time, a ready-to-use patch-level retrieval framework across H\&E, mIF, and text modalities. Beyond retrieval, the learned representations encode rich tissue context and enable accurate downstream classification of tissue type, tumor status, and challenging tumor stage labels, consistently outperforming unimodal baselines. These results indicate that Haiku not only aligns heterogeneous modalities, but also captures the shared biological and pathological semantics underlying them. Importantly, tri-modal alignment allows the model to leverage complementary information across modalities that cannot be recovered from any single modality alone.

We further introduce two downstream applications that leverage the tri-modal representation. For biomarker inference, combining H\&E embeddings with metadata-only text descriptions significantly improves prediction accuracy over H\&E-only retrieval, providing direct evidence that tri-modal alignment transfers complementary semantic knowledge into molecular prediction. Our experiments further show that naively converting multiplexed protein channels into three-channel RGB images fails to effectively leverage H\&E--pretrained models, underscoring the necessity of different image modality-specific foundation models. Notably, several recent studies have attempted to predict molecular biomarkers directly from H\&E images~\cite{Wu2025-nj, Valanarasu2025-ek}. However, these approaches typically frame the problem as pixel-level segmentation or classification and are trained as discriminative models. In contrast, Haiku emphasizes learning a unified multimodal representation rather than a task-specific predictor. Crucially, our retrieval-based design is \emph{evidence-based}: every predicted biomarker pattern is grounded in real mIF patches from the held-out gallery that can be visually inspected, rather than synthesized de novo. Unlike purely generative approaches, which can hallucinate plausible-looking but unverifiable outputs, retrieval guarantees that the returned molecular evidence always corresponds to actually measured tissue. Our zero-shot fusion retrieval results further demonstrate that metadata-conditioned inference, enabled by tri-modal alignment, outperforms unimodal H\&E-based approaches, highlighting the necessity of incorporating complementary textual semantics for accurate biomarker prediction. This representation-centric design enables a broader spectrum of downstream uses, including retrieval, cross-modal reasoning, and counterfactual analysis. In this sense, Haiku serves as a general connective framework that bridges modalities and facilitates exploratory analyses across clinical imaging and spatial molecular profiling.

For counterfactual analysis, we apply two complementary perturbation paradigms: staging-based perturbation in breast cancer to characterize progression dynamics, and survival-based perturbation in lung cancer to explore molecular signals associated with patient outcome. In both cases, modifying only clinical metadata while fixing H\&E morphology surfaces niche-specific shifts in retrieved mIF biomarker abundance, which we present as exploratory rather than mechanistic findings. The cluster-stratified heatmap (\textbf{Figure~\ref{fig:figure5}g}) shows that these shifts are localized to specific morphological compartments rather than being uniformly distributed, for example, with counterfactual myeloid enrichment and epithelial-lineage loss most pronounced in the \emph{epithelial-dominant tumor core} (C3) while the \emph{fibroblast-rich stroma} (C0) compartment shows distinct shifts, suggesting that morphological stratification is useful for interpreting such results. Combining the cluster-stratified shifts, the cross-compartment magnitude gradient, and the within-compartment baseline-conditioned response, Haiku's counterfactual framework reveals biomarker changes that are both compartment-specific and patch-conditioned from a metadata-only perturbation, with directions consistent with canonical features of advanced-stage breast cancer biology in the appropriate morphological niches. The same retrieval-based, evidence-based design carries over to this setting: counterfactual outcomes are defined through differences between two retrieved mIF patch sets that can be visually inspected, rather than through direct prediction of continuous biomarker values. Users can define their own evidence by changing only the target mIF reference set without retraining, supporting exploration of diverse clinical and biological questions through a single pretrained model. We emphasize, however, that broader cohorts and orthogonal validation will be needed to assess the generalizability of any specific recovered pattern beyond the single-patient case studies presented here.

More broadly, our results support the notion that integrating heterogeneous modalities yields benefits beyond incremental performance improvements. Well-aligned multimodal embeddings provide a principled mechanism to disentangle shared biological signals from modality-specific noise, leading to more robust and generalizable models. This observation resonates with the Platonic Representation Hypothesis~\cite{Huh2024-wy}, which suggests that large-scale models trained across diverse views of the same system converge toward a common latent representation. We hypothesize that such convergence may extend to pathology, where H\&E, spatial proteomics, text, and other spatial omics modalities share an underlying biological structure that can be captured within a unified latent space~\cite{Guo2025-jj}.

Despite these strengths, Haiku has several limitations. First, the current model is trained on paired datasets only; incorporating mixtures of paired and unpaired data could further improve scalability and enable more efficient utilization of large unimodal corpora~\cite{Wang2024-hs}. Second, contrastive alignment performance depends on the quality of modality-specific encoders. While this dependency may constrain current performance, it also allows Haiku to directly benefit from future advances in pretrained encoders. Third, text descriptions are generated from structured metadata templates rather than free-text clinical narratives; extending Haiku to handle heterogeneous real-world clinical text remains an open challenge. Fourth, the counterfactual analyses presented here are single-patient proof-of-concept demonstrations (281 patches from one T2N0 breast cancer case and 154 patches from one lung adenocarcinoma case); population-level validation across larger patient cohorts and experimental confirmation of the inferred molecular mechanisms would further strengthen the biological conclusions. Finally, the current framework operates at the patch level ($256 \times 256$ pixels); integration with whole-slide-level architectures can enable broader clinical deployment.

Overall, Haiku provides both a multimodal pretrained model and a conceptual framework for unifying clinical imaging and molecular profiling. By bridging spatial proteomics, histology, and textual semantics within a shared representation space, Haiku establishes a foundation for multimodal computational pathology and opens new opportunities for integrative biological discovery and translational research.

\section{Methods}
\label{sec:methods}
\subsection{Dataset introduction}

Our dataset is drawn from a multi-center, multi-disease cohort curated by Enable Medicine and comprises 7{,}600 mIF tissue slices in total, of which 3{,}554 are paired with co-registered H\&E and clinical metadata and 4{,}046 are mIF-only. For held-out evaluation, we randomly select 336 slices from the paired subset and 198 slices from the mIF-only subset, yielding 534 held-out test slices in total (\textbf{Figure~\ref{fig:figure1}b}). The remaining 7{,}066 slices (3{,}218 paired $+$ 3{,}848 unpaired) are used for training: paired slices support Haiku tri-modal contrastive alignment, while the full 7{,}066-slice training pool is used to pretrain the VirTues mIF encoder. The 534 held-out slices are excluded from both VirTues pretraining and Haiku contrastive training, enforcing a strict separation between training and evaluation data and preventing any form of data leakage. The train/test partition is performed at the patient level: of the $1{,}848$ unique patients in the cohort, $1{,}606$ ($86.9\%$) are assigned to the training pool and $242$ ($13.1\%$) are reserved for held-out testing, so that all tissue slices and patches originating from a given patient remain within a single partition, preventing patient-level leakage between training and evaluation across all downstream protocols (\textbf{Supplementary Figure~\ref{supplementary:2}}).

\subsection{Data preprocessing}
\label{sec:datapreprocess}
To enable aligned tri-modal representation learning, all three modalities---H\&E images, mIF multiplexed protein images, and text descriptions---are processed at a matched patch level. For the two image modalities, H\&E and mIF images corresponding to the same tissue slice are pre-registered and share a common coordinate system. As a result, patch coordinates derived from one modality can be directly applied to the other, ensuring that each H\&E patch, mIF patch, and text description corresponds to the same tissue region and forms a valid positive sample for contrastive learning.

\subsubsection{mIF image normalization and patch extraction}

For each tissue region $r$, mIF data are provided as a multi-channel image
\[
I^{(r)}_{\mathrm{mIF}} \in \mathbb{R}^{H_r \times W_r \times C_r},
\]
where $C_r$ denotes the number of biomarker channels available for region $r$. A corresponding binary tissue mask is obtained using a pretrained ResNet50~\cite{he2016deep}-based segmentation network with the DAPI channel as input:
\[
\mathbf{M}^{(r)} \in \{0,1\}^{H_r \times W_r},
\]
where $\mathbf{M}^{(r)}=1$ denotes foreground tissue and $\mathbf{M}^{(r)}=0$ denotes background.

\paragraph{Channel-wise normalization.}
Each mIF biomarker channel is independently normalized to reduce staining variability and suppress background noise. For a given tissue region $r$ and biomarker channel $c$, we consider the single-channel image $I^{(r)}_{c} \in \mathbb{R}^{H_r \times W_r}$ together with the binary tissue mask $\mathbf{M}^{(r)} \in \{0,1\}^{H_r \times W_r}$.

Background pixels ($\mathbf{M}^{(r)}=0$) are used to estimate a lower intensity bound as the median background signal, while foreground pixels ($\mathbf{M}^{(r)}=1$) are used to estimate an upper bound based on the $99$th percentile of foreground intensities, scaled by a factor of $1.1$ to preserve high-signal structures. To avoid degenerate normalization ranges, the upper bound is further constrained to be at least a fixed margin above the background level and no greater than the maximum 16-bit intensity value.

Given these provisional bounds, we apply an adaptive histogram-based thresholding procedure to refine the effective clipping range. Specifically, a histogram of pixel intensities is computed within the provisional range, and only bins whose counts fall within a predefined frequency interval are retained. The lowest and highest retained bins define the final lower and upper clipping thresholds for the channel. If no valid bins are identified, the provisional bounds are used as a fallback.

Pixel intensities are then clipped to the final range and linearly rescaled to $[0,1]$, followed by quantization to 8-bit resolution. This procedure is applied independently to each biomarker channel, yielding a normalized multi-channel mIF image $\hat{I}^{(r)}_{\mathrm{mIF}} \in \mathbb{R}^{C_r \times H_r \times W_r}$, which is subsequently used for patch extraction and downstream modeling.

\paragraph{Sliding-window patch generation.}
Normalized mIF images are decomposed into square patches of size
\[
P \times P \quad \text{with } P = 256.
\]
Patches are generated using a sliding-window strategy with stride
\[
S = \lfloor 0.7P \rfloor,
\]
and random spatial jitter is applied to the window origin to reduce grid artifacts.

Only patches with tissue coverage greater than $90\%$, as determined by $\mathbf{M}^{(r)}$, are retained. For each retained patch $k$ in region $r$, all biomarker channels are cropped at identical spatial coordinates
\[
\mathrm{Coord}^{(r)}_k = (x_{\mathrm{left}}, x_{\mathrm{right}}, y_{\mathrm{bottom}}, y_{\mathrm{top}}),
\]
yielding the mIF patch
\[
p^{(r)}_{k,\mathrm{mIF}} =
\hat{I}^{(r)}_{\mathrm{mIF}}
[y_{\mathrm{bottom}}:y_{\mathrm{top}},\, x_{\mathrm{left}}:x_{\mathrm{right}}].
\]
These coordinates are reused to extract paired H\&E patches.

\subsubsection{H\&E image patch extraction}

Whole-slide H\&E images are provided as high-resolution RGB images
\[
I^{(r)}_{\mathrm{HE}} \in \mathbb{R}^{H_r \times W_r \times 3}.
\]
For each patch $k$ in region $r$, the corresponding H\&E patch is extracted using the same coordinates:
\[
p^{(r)}_{k,\mathrm{HE}} =
I^{(r)}_{\mathrm{HE}}
[y_{\mathrm{bottom}}:y_{\mathrm{top}},\, x_{\mathrm{left}}:x_{\mathrm{right}}],
\]
yielding spatially aligned patch pairs
\[
p^{(r)}_{k,\mathrm{HE}} \in \mathbb{R}^{P \times P \times 3},
\quad
p^{(r)}_{k,\mathrm{mIF}} \in \mathbb{R}^{P \times P \times C_r}.
\]

\subsubsection{Patch-level text description generation}
\label{sec:textdescript}

For each mIF patch, we generate a structured natural language description summarizing local biomarker expression, spatial organization, and available clinical metadata. This textual modality provides a compact yet semantically rich representation of the molecular and spatial context of each patch.

\paragraph{Patch-level biomarker quantification.}
Let $p^{(r)}_{k,\mathrm{mIF}}$ denote the $k$-th mIF patch from region $r$. For each biomarker channel $c \in \{1,\dots,C_r\}$, we compute the mean foreground intensity
\[
\mu^{(r)}_{k,c}
=
\frac{1}{|\Omega^{(r)+}_{k,c}|}
\sum_{(x,y)\in\Omega^{(r)+}_{k,c}}
\hat{I}^{(r)}_{k,c}(x,y),
\]
where $\Omega^{(r)+}_{k,c}$ denotes foreground pixels with non-zero signal.

To contextualize expression within the slice, we compute region-normalized statistics using all other patches:
\[
z^{(r)}_{k,c}
=
\frac{\mu^{(r)}_{k,c} - \bar{\mu}^{(r)}_{-k,c}}
{\sigma^{(r)}_{-k,c} + \epsilon},
\qquad
\pi^{(r)}_{k,c}
=
\frac{1}{N_r-1}
\sum_{j\neq k}
\mathbb{I}(\mu^{(r)}_{j,c} \le \mu^{(r)}_{k,c}),
\]
where $N_r$ is the number of patches in region $r$.

\paragraph{Spatial distribution characterization.}
For each single-channel mIF patch $p^{(r)}_{k,c} \in \mathbb{R}^{P \times P}$, pixel intensities are first rescaled to an 8-bit range for texture analysis, and foreground pixels are identified as those with non-zero signal. Let $\Omega^{(r)+}_{k,c}$ denote the set of foreground pixels.

We compute a set of complementary spatial statistics capturing signal variability, organization, and coverage. Signal heterogeneity is quantified by the coefficient of variation,
\[
\mathrm{CV}^{(r)}_{k,c}
=
\frac{\sigma^{(r)}_{k,c}}{\mu^{(r)}_{k,c} + \epsilon},
\]
where $\mu^{(r)}_{k,c}$ and $\sigma^{(r)}_{k,c}$ are the mean and standard deviation of foreground intensities, respectively. Local spatial organization is characterized using gradient-based clustering, defined as the inverse of the mean gradient magnitude over foreground pixels,
\[
\mathrm{Clust}^{(r)}_{k,c}
=
\frac{1}{1 + \mathbb{E}_{(x,y)\in\Omega^{(r)+}_{k,c}}
\left[\sqrt{(\nabla_x p^{(r)}_{k,c})^2 + (\nabla_y p^{(r)}_{k,c})^2}\right]},
\]
such that higher values correspond to more spatially clustered signal.

In addition, second-order texture features are extracted from the gray-level co-occurrence matrix (GLCM) computed with pixel distance $1$ and angle $0$, including homogeneity and contrast, which capture local smoothness and intensity transitions. Signal coverage is defined as
\[
\mathrm{Cov}^{(r)}_{k,c}
=
\frac{|\Omega^{(r)+}_{k,c}|}{P^2},
\]
representing the fraction of patch area occupied by foreground signal.

\paragraph{Rule-based spatial pattern assignment.}
The continuous spatial metrics are mapped to discrete spatial pattern labels using fixed, interpretable rules based on coverage, variability, and clustering strength. Specifically,
\[
\mathcal{D}^{(r)}_{k,c} \in
\{\text{sparse}, \text{uniform}, \text{clustered}, \text{heterogeneous}\},
\]
with subcategories reflecting signal density.

Patches with low coverage ($\mathrm{Cov}<0.1$) are labeled as \emph{sparse}.  
For high-coverage patches ($\mathrm{Cov}>0.7$), patterns are assigned as \emph{uniform} if signal variability is low ($\mathrm{CV}<0.5$) and texture homogeneity is high ($>0.6$), as \emph{clustered} if spatial clustering is strong ($\mathrm{Clust}>0.7$), and as \emph{heterogeneous} otherwise.  
For intermediate coverage ($0.3<\mathrm{Cov}\le0.7$), patches are labeled as \emph{clustered} when clustering exceeds $0.6$, as \emph{uniform} when variability is low ($\mathrm{CV}<0.6$), and as \emph{heterogeneous} otherwise.  
Remaining low-coverage patches are classified as \emph{clustered} or \emph{scattered} based on whether clustering exceeds $0.5$.

These discrete spatial descriptors are subsequently used in patch-level text synthesis to provide interpretable summaries of biomarker spatial organization.

\paragraph{Clinical metadata integration and text synthesis.}
Region-level metadata $\mathcal{M}_r$, including tissue type, disease status, and experimental annotations, are incorporated when available. The final patch-level description is generated using template-based synthesis:
\[
p^{(r)}_{k,\mathrm{TXT}}
=
\mathrm{TextGen}
\big(
\mathcal{S}^{(r)}_k,\,
\{\mathcal{D}^{(r)}_{k,c}\},\,
\mathcal{M}_r
\big),
\]
where $\mathcal{S}^{(r)}_k = \{(c, z^{(r)}_{k,c}, \pi^{(r)}_{k,c})\}$ denotes the biomarker summary. Numerical values are used internally for ranking but omitted from the final text.

\subsection{Overview of the tri-modal representation learning framework}

We develop a tri-modal representation learning framework that jointly embeds histology images, multiplexed protein images, and text descriptions into a shared latent space. Each modality is processed by a dedicated encoder followed by a modality-specific projection head. All components are trained using contrastive pretraining to enforce cross-modal alignment.

Let
\[
\mathcal{D}
=
\{(x_i^{\mathrm{HE}}, x_i^{\mathrm{mIF}}, x_i^{\mathrm{TXT}})\}_{i=1}^{N}
\]
denote the training dataset, where $x_i^{\mathrm{HE}}$ is an H\&E image patch, $x_i^{\mathrm{mIF}}$ is the corresponding multiplexed protein (mIF) image patch, and $x_i^{\mathrm{TXT}}$ is the associated textual description derived from molecular, spatial, and clinical attributes.

More specifically, let $r \in \mathcal{R}$ index tissue regions, and let each region $r$ contain $K^{(r)}$ patches. The total number of patches satisfies
\[
N = \sum_{r \in \mathcal{R}} K^{(r)}.
\]
For each patch index $i$, we denote by $r_i$ the corresponding region and by $k_i$ the patch index within that region. Accordingly,
\[
x_i^{\mathrm{mIF}} = p^{(r_i)}_{k_i,\mathrm{mIF}}, \quad
x_i^{\mathrm{HE}} = p^{(r_i)}_{k_i,\mathrm{HE}}, \quad
x_i^{\mathrm{TXT}} = p^{(r_i)}_{k_i,\mathrm{TXT}},
\]
where all three modalities correspond to the same spatial tissue location.

Each modality is encoded using a modality-specific pretrained encoder:
\[
\mathbf{h}_i^{\mathrm{HE}} = f_{\mathrm{HE}}(x_i^{\mathrm{HE}}), \quad
\mathbf{h}_i^{\mathrm{mIF}} = f_{\mathrm{mIF}}(x_i^{\mathrm{mIF}}), \quad
\mathbf{h}_i^{\mathrm{TXT}} = f_{\mathrm{TXT}}(x_i^{\mathrm{TXT}}).
\]

For the H\&E modality, we adopt the pretrained MUSK vision transformer as the image encoder. For the mIF modality, as no large-scale publicly available pretrained models exist for multiplexed protein images, we pretrain a VirTues encoder from scratch using our private mIF dataset, which includes all available mIF-only slices as well as the training portion of paired slices used for tri-modal alignment. For the text modality, we employ BiomedBERT~\cite{Gu2021-biomedbert}, a BERT encoder pretrained from scratch on PubMed abstracts and PubMedCentral full-text articles.

In addition, to address the irregularity of biomarker channels in mIF data, we follow the VirTues strategy of encoding each biomarker using its corresponding ESM-3~\cite{Hayes2025-gn} protein embedding, matched to the protein identity of each biomarker. This design enables the model to accept biomarkers that do not appear in the pretraining set, provided that a corresponding ESM embedding is available. For biomarkers such as DAPI that do not have a corresponding protein embedding, we initialize a learnable embedding during VirTues pretraining and fix this embedding during inference.

\subsubsection{Projection heads}

Each modality-specific encoder output is mapped into a shared latent space using a projection head:
\[
\mathbf{z}_i^{m} = g_m(\mathbf{h}_i^{m}),
\quad
m \in \{\mathrm{HE}, \mathrm{mIF}, \mathrm{TXT}\}.
\]

All projection heads share the same architecture but maintain independent parameters across modalities. Each projection head is implemented as a two-layer multilayer perceptron with output dimension $d = 512$:
\[
g_m(\mathbf{h})
=
\mathrm{BN}
\big(
W_2 \, \mathrm{ReLU}(W_1 \mathbf{h})
\big).
\]

All projected embeddings are $\ell_2$-normalized prior to contrastive learning:
\[
\tilde{\mathbf{z}}_i^{m}
=
\frac{\mathbf{z}_i^{m}}{\lVert \mathbf{z}_i^{m} \rVert_2}.
\]

\subsubsection{Tri-modal contrastive pretraining}
\label{sec:training}

\paragraph{Training objective}

For a minibatch of size $B = 128$, the model produces normalized embeddings
\[
\{\tilde{\mathbf{z}}_i^{\mathrm{HE}}, \tilde{\mathbf{z}}_i^{\mathrm{mIF}}, \tilde{\mathbf{z}}_i^{\mathrm{TXT}}\}_{i=1}^{B}.
\]

For any ordered modality pair $(a,b)$, we define a contrastive loss
\[
\mathcal{L}_{a \rightarrow b}
=
-\frac{1}{B}
\sum_{i=1}^{B}
\log
\frac{
\exp \left(
\tilde{\mathbf{z}}_i^{a} \cdot \tilde{\mathbf{z}}_i^{b} / \tau
\right)
}{
\sum_{j=1}^{B}
\exp \left(
\tilde{\mathbf{z}}_i^{a} \cdot \tilde{\mathbf{z}}_j^{b} / \tau
\right)
},
\]
where similarities are computed using cosine similarity in the shared latent space.

The temperature parameter is fixed to $\tau = 0.07$. Losses are computed for all cross-modal pairs and summed to obtain the final training objective:
\[
\mathcal{L}
=
\mathcal{L}_{\mathrm{HE}\rightarrow\mathrm{mIF}}
+
\mathcal{L}_{\mathrm{mIF}\rightarrow\mathrm{TXT}}
+
\mathcal{L}_{\mathrm{HE}\rightarrow\mathrm{TXT}},
\]
with symmetric counterparts implicitly included.

\paragraph{Optimization and training details}

Models are trained using the AdamW optimizer. During tri-modal alignment, the H\&E and text encoders are fine-tuned in the final two transformer blocks, while the mIF encoder is kept frozen. Distinct learning rates are applied to different parameter groups: $1 \times 10^{-5}$ for the H\&E encoder, $2 \times 10^{-5}$ for the text encoder, and $1 \times 10^{-4}$ for all projection heads.

A two-stage learning rate schedule is employed. Learning rates are linearly warmed up over the first 5,000 optimization steps, followed by cosine annealing for the remainder of training. The model is trained for 25 epochs over the full training dataset.

\subsection{Downstream evaluation}

To comprehensively evaluate the representations learned by Haiku, we assess performance across a diverse set of downstream tasks spanning patch-level retrieval, classification, biomarker inference, and slice-level clinical prediction. Unless otherwise specified, all downstream evaluations use frozen pretrained encoders without fine-tuning backbone parameters. 
 
For baseline comparisons, we extract unimodal embeddings from MUSK (H\&E) and VirTues (mIF) using the backbone outputs after our tri-modal alignment training (i.e., prior to the Haiku projection heads). Throughout the paper, MUSK and VirTues baselines refer to these identically fine-tuned backbones, evaluated under the same splits and downstream protocols as Haiku to ensure a fair comparison.

\subsubsection{Cross-modality patch-level retrieval}
\label{sec:retrieval}

We evaluate cross-modality retrieval to assess whether Haiku learns a shared latent space in which semantically corresponding tissue patches from different modalities are consistently aligned. In this task, each query corresponds to a single patch represented in one modality
\[
a \in \{\mathrm{HE}, \mathrm{mIF}, \mathrm{TXT}\},
\]
and the objective is to retrieve relevant patches from a target modality
\[
b \in \{\mathrm{HE}, \mathrm{mIF}, \mathrm{TXT}\}, \quad b \neq a,
\]
based on similarity in the learned embedding space.

Let $\mathcal{H}$ denote the held-out evaluation set consisting of $N$ spatially aligned patches across all modalities. For each patch $i \in \mathcal{H}$, we obtain modality-specific normalized embeddings
\[
\tilde{\mathbf{z}}^{a}_i, \quad \tilde{\mathbf{z}}^{b}_i,
\]
where $a$ denotes the query modality and $b$ denotes the target modality.

\paragraph{Query and gallery construction.}
For a given retrieval direction $a \rightarrow b$, the query set is defined as
\[
\mathcal{Q}^{a} = \{\tilde{\mathbf{z}}^{a}_i\}_{i=1}^{N},
\]
and the gallery set is defined as
\[
\mathcal{G}^{b} = \{\tilde{\mathbf{z}}^{b}_j\}_{j=1}^{N},
\]
where the gallery contains \emph{all} patches from the target modality in the held-out set. Thus, the number of query and gallery embeddings is identical, and retrieval is performed globally across the entire held-out dataset rather than within individual tissue slices.

\paragraph{Similarity and ranking.}
For each query patch $i \in \mathcal{Q}^{a}$, cosine similarity is computed against all gallery patches:
\[
s_{ij}
=
\tilde{\mathbf{z}}^{a}_i \cdot \tilde{\mathbf{z}}^{b}_j,
\qquad
\tilde{\mathbf{z}} = \frac{\mathbf{z}}{\lVert \mathbf{z}\rVert_2}.
\]
Gallery items are ranked in descending order of similarity, yielding a ranked list
\[
\pi_i = \mathrm{argsort}_{j \in \mathcal{G}^{b}}(s_{ij}).
\]

\paragraph{Relevance definition.}
Relevance between a query $i$ and a gallery item $j$ is encoded by a binary indicator $y_{ij} \in \{0,1\}$. In the primary evaluation setting, relevance is defined by spatial correspondence:
\[
y_{ij} = \mathbb{I}(i=j),
\]
i.e., the ground-truth match for each query patch is the gallery patch originating from the same spatial location in the same tissue section.

\subsubsection{Evaluation metrics}

Retrieval performance is quantified using top-$K$ metrics derived from the ranked gallery list $\pi_i$ for each query.

Top-$K$ accuracy measures whether the correct gallery patch appears within the first $K$ retrieved results:
\[
\mathrm{Recall@}K(i)
=
\mathbb{I}\!\left(
\sum_{t=1}^{K} y_{i,\pi_i(t)} > 0
\right).
\]
The reported score is the macro-average across all queries:
\[
\mathrm{Recall@}K
=
\frac{1}{N}
\sum_{i=1}^{N} \mathrm{Recall@}K(i).
\]

All retrieval evaluations are performed on held-out tissue sections that are not used during contrastive pretraining. For each retrieval direction, the query and gallery sets span all patches in the held-out dataset, resulting in a large-scale cross-sample retrieval setting that closely reflects real-world deployment scenarios.

\subsubsection{K-nearest-neighbor (KNN) classification evaluation}
\label{sec:knn}

In addition to retrieval-based evaluation, we assess representation quality using a K-nearest-neighbor (KNN) classification protocol. While cross-modality retrieval evaluates whether the correct paired patch appears among the top-ranked candidates, KNN classification explicitly tests whether local neighborhoods in the shared embedding space are label-consistent.

\paragraph{Relationship to retrieval evaluation.}
KNN classification uses the same query and gallery embeddings as the retrieval task, where the gallery consists of all patch embeddings in the target modality from the held-out set. However, instead of checking whether a specific paired patch is retrieved, KNN classification predicts a semantic label for each query by aggregating labels from its top-$K$ nearest neighbors in the gallery.

\paragraph{KNN prediction rule.}
For a query embedding $\mathbf{z}_i^{a}$ in modality $a$, cosine similarity is computed against all gallery embeddings $\{\mathbf{z}_j^{b}\}_{j=1}^{N}$ in modality $b$. Let $\mathcal{N}_K(i)$ denote the indices of the top-$K$ most similar gallery embeddings.

Each gallery embedding $\mathbf{z}_j^{b}$ is associated with a categorical label $y_j \in \mathcal{Y}$. The predicted label $\hat{y}_i^{(K)}$ for query $i$ is obtained by similarity-weighted voting:
\[
\hat{y}_i^{(K)}
=
\arg\max_{c \in \mathcal{Y}}
\sum_{j \in \mathcal{N}_K(i)}
\mathbb{I}(y_j = c)\;
\mathrm{sim}(\mathbf{z}_i^{a}, \mathbf{z}_j^{b}),
\]
where $\mathrm{sim}(\cdot,\cdot)$ denotes cosine similarity and $\mathbb{I}(\cdot)$ is the indicator function.

All embeddings are $\ell_2$-normalized prior to similarity computation. In this work, we primarily evaluate KNN classification performance using the 1-nearest-neighbor setting ($K=1$), unless otherwise specified.

Classification performance is quantified using macro-averaged F1 score (F1$_\mathrm{macro}$@K). All metrics are computed by comparing predicted labels $\hat{y}_i^{(K)}$ against ground-truth query labels $y_i$ and are averaged across all queries in the held-out evaluation set.

\paragraph{Macro-averaged F1 score (F1$_\mathrm{macro}$@K).}
For each class $c \in \mathcal{Y}$, the class-wise F1 score is defined as
\[
\mathrm{F1}_c
=
\frac{
2 \cdot \mathrm{TP}_c
}{
2 \cdot \mathrm{TP}_c + \mathrm{FP}_c + \mathrm{FN}_c
}.
\]
The macro-averaged F1 score is obtained by averaging over all classes:
\[
\mathrm{F1}_{\mathrm{macro}}@K
=
\frac{1}{|\mathcal{Y}|}
\sum_{c \in \mathcal{Y}}
\mathrm{F1}_c.
\]

\subsubsection{Zero-shot cross-modality classification}
\label{sec:zeroshot}

\paragraph{Task definition and dataset construction.}
We evaluate zero-shot classification to assess whether Haiku aligns mIF embeddings with semantic class descriptions in the shared latent space. For each classification task (e.g., disease category or tissue type), we use a held-out evaluation set of $N$ image patches with integer class labels
\[
y_i \in \{0,1,\ldots,C-1\}, \qquad i=1,\ldots,N,
\]
where $C$ denotes the number of classes. For each patch, we use frozen image embeddings from mIF,
\[
\{\mathbf{z}^{\mathrm{mIF}}_i\}_{i=1}^{N},
\]
and perform classification without training any additional classifier.

For the text side, we generate one class description per class using prompt templates. Concretely, for each class name $\ell_c$ (the $c$-th class) and each template $t(\cdot)$, we form a class prompt
\[
x^{\mathrm{TXT}}_{c,t} = t(\ell_c),
\qquad c=0,\ldots,C-1.
\]
We use multiple prompt templates to reduce template sensitivity and obtain more stable estimates. In our experiments, we evaluate five templates per task (e.g., ``A mIF region of $\{\}$ disease.''), and report the mean and standard deviation of metrics across templates.

\paragraph{Prediction rule.}
For a fixed template $t$, we encode all class prompts using the text encoder to obtain a class text embedding matrix
\[
\mathbf{T}^{(t)}
=
\begin{bmatrix}
\mathbf{z}^{\mathrm{TXT}}_{0,t}\\
\mathbf{z}^{\mathrm{TXT}}_{1,t}\\
\vdots\\
\mathbf{z}^{\mathrm{TXT}}_{C-1,t}
\end{bmatrix}
\in \mathbb{R}^{C \times d}.
\]
All mIF and text embeddings are $\ell_2$-normalized before similarity computation. Given an mIF embedding $\tilde{\mathbf{z}}^{\mathrm{mIF}}_i$, we compute cosine similarities to all class text embeddings:
\[
\mathbf{s}^{(t)}_i
=
\tilde{\mathbf{z}}^{\mathrm{mIF}}_i \, \big(\tilde{\mathbf{T}}^{(t)}\big)^{\!\top}
\in \mathbb{R}^{C},
\]
and predict the class by nearest text prototype:
\[
\hat{y}^{(t)}_i
=
\arg\max_{c \in \{0,\ldots,C-1\}}
\mathbf{s}^{(t)}_{i,c}.
\]

\paragraph{Evaluation metrics.}
For each template $t$, we compute macro-averaged F1 by comparing predictions $\hat{y}^{(t)}_i$ to ground-truth labels $y_i$:

\[
\mathrm{F1}^{(t)}_{\mathrm{macro}}
=
\frac{1}{C}\sum_{c=0}^{C-1}
\frac{2\,\mathrm{TP}_c^{(t)}}{2\,\mathrm{TP}_c^{(t)}+\mathrm{FP}_c^{(t)}+\mathrm{FN}_c^{(t)}}.
\]

Final reported metrics are summarized across prompt templates by reporting the mean and standard deviation.

\paragraph{Random-guess baseline.}
To contextualize performance, we additionally report a random-guess baseline by sampling predicted labels from a uniform distribution over classes and evaluating the same metrics, repeating this procedure multiple times (10 times in our experiments) to estimate the mean and standard deviation.

\subsubsection{Patch-level linear probing classification}
\label{sec:linearprob}

\paragraph{Task definition and dataset construction.}
To assess representation quality under supervised adaptation, we perform patch-level linear probing on frozen embeddings derived from different modalities. Patch-level labels curated from clinical metadata include organ type, tissue type, tumor T stage, tumor N stage, and tumor grade (G1--G3). Tumor-related tasks are restricted to cancer samples.

To prevent spatial leakage, five-fold cross-validation is performed at the tissue slice level, ensuring that patches originating from the same slice do not appear in both training and test sets.

\paragraph{Evaluated representations.}
We evaluate linear probing performance using the following patch-level representations:
\begin{itemize}
\item \textbf{Haiku(H\&E)}: embeddings extracted from the H\&E encoder output after the projection head,
\item \textbf{Haiku(mIF)}: embeddings extracted from the mIF encoder output after the projection head,
\item \textbf{Haiku(Fusion)}: concatenated embeddings formed by channel-wise concatenation of \textbf{Haiku(H\&E)} and \textbf{Haiku(mIF)} embeddings,
\[
\mathbf{z}_i^{\mathrm{fusion}}
=
\big[
\mathbf{z}_i^{\mathrm{HE}} \,\Vert\, \mathbf{z}_i^{\mathrm{mIF}}
\big],
\]
where $\Vert$ denotes vector concatenation.
\end{itemize}

This fusion strategy is designed to test whether complementary information from the two imaging modalities improves linear separability.

\paragraph{Linear probing model.}
For each representation, we train a multinomial logistic regression classifier on top of fixed embeddings. Given embeddings $\{\mathbf{z}_i\}_{i=1}^{N}$ and labels $\{y_i\}_{i=1}^{N}$, the classifier models class probabilities as
\[
p(y=c \mid \mathbf{z})
=
\frac{\exp(\mathbf{w}_c^\top \mathbf{z} + b_c)}
     {\sum_{c'} \exp(\mathbf{w}_{c'}^\top \mathbf{z} + b_{c'})}.
\]
The classifier is optimized using cross-entropy loss with $\ell_2$ regularization.

\paragraph{Hyperparameter selection.}
To ensure fair comparison across modalities and baselines, we apply an identical hyperparameter selection protocol to all models. Specifically, we perform grid search over the regularization parameter
\[
C \in \{0.1, 1, 10\},
\]
using the same five-fold stratified cross-validation splits. For each task and representation, the optimal value $C^\ast$ is selected based on the mean macro-averaged F1 score across folds. All reported results correspond to performance achieved using $C^\ast$.

\paragraph{Baselines and fair comparison.}
To ensure a fair and controlled comparison, we include unimodal baselines based on MUSK and VirTues, as well as a naive majority-vote baseline, which predicts the most frequent class in the training split for all test samples within each fold.

Both VirTues and MUSK baselines are evaluated using the same cross-validation splits, classifier architecture, and hyperparameter grid, ensuring that performance differences arise from representation quality rather than optimization advantages.

\paragraph{Evaluation metrics.}
For each fold, we evaluate performance using the macro-averaged F1 score (F1$_\mathrm{macro}$) to account for class imbalance in the dataset. Final results are reported as the mean and standard deviation of the metric across the five cross-validation folds.

\subsubsection{Slice-level prediction using multiple-instance learning}
\label{sec:mil}

\paragraph{Task definition and bag construction.}
For slice-level clinical prediction, we adopt a multiple-instance learning (MIL) formulation in which each slice is represented as a bag of mIF patch embeddings. Let $\mathcal{I}$ denote the set of evaluation slices. For each slice $i\in\mathcal{I}$, we define the bag
\[
\mathcal{B}_i
=
\left\{\mathbf{z}^{\mathrm{mIF}}_{i1},\mathbf{z}^{\mathrm{mIF}}_{i2},\ldots,\mathbf{z}^{\mathrm{mIF}}_{i n_i}\right\},
\quad
\mathbf{z}^{\mathrm{mIF}}_{ij}\in\mathbb{R}^{D},
\]
where $n_i$ is the number of available patches for slice $i$. All instance embeddings are $\ell_2$-normalized prior to MIL modeling:
\[
\tilde{\mathbf{z}}^{\mathrm{mIF}}_{ij}
=
\frac{\mathbf{z}^{\mathrm{mIF}}_{ij}}{\left\lVert \mathbf{z}^{\mathrm{mIF}}_{ij}\right\rVert_2}.
\]
During minibatch training, variable-length bags are padded to a common length and accompanied by a binary mask $\mathbf{m}_i\in\{0,1\}^{n_i}$ indicating valid instances.

\paragraph{Attention-based MIL pooling.}
Given a padded instance matrix $\mathbf{X}_i\in\mathbb{R}^{n_i\times D}$ for slice $i$, we first compute instance-level hidden features using an instance encoder $\phi(\cdot)$ applied independently to each instance:
\[
\mathbf{H}_i=\phi(\mathbf{X}_i)\in\mathbb{R}^{n_i\times d},
\quad
\mathbf{h}_{ij}\in\mathbb{R}^{d}.
\]
Attention weights over instances are then computed as
\[
a_{ij}
=
\frac{
\exp\!\left(\mathbf{w}^\top \tanh(\mathbf{V}\mathbf{h}_{ij})\right)
}{
\sum\limits_{j':\, m_{ij'}=1}
\exp\!\left(\mathbf{w}^\top \tanh(\mathbf{V}\mathbf{h}_{ij'})\right)
},
\]
where $\mathbf{V}\in\mathbb{R}^{d_a\times d}$ and $\mathbf{w}\in\mathbb{R}^{d_a}$ are learnable parameters, and padded instances ($m_{ij}=0$) are excluded from the normalization. The slice-level representation is obtained as an attention-weighted sum:
\[
\mathbf{s}_i
=
\sum_{j:\,m_{ij}=1} a_{ij}\mathbf{h}_{ij}
\in\mathbb{R}^{d}.
\]
This pooled embedding $\mathbf{s}_i$ serves as the input to task-specific prediction heads.

\subsubsection{MIL classification for treatment response and clinical endpoints}

\paragraph{Prediction head and inference.}
For binary classification tasks, each slice $i$ is associated with a label $y_i\in\{0,1\}$. A classification head $g_{\mathrm{cls}}(\cdot)$ maps the pooled representation to a scalar logit:
\[
\hat{\ell}_i=g_{\mathrm{cls}}(\mathbf{s}_i)\in\mathbb{R},
\qquad
\hat{p}_i=\sigma(\hat{\ell}_i),
\]
where $\sigma(\cdot)$ denotes the sigmoid function.

\paragraph{Training objective.}
Model parameters are optimized using a positive-class weighted binary cross-entropy loss to mitigate class imbalance.

\paragraph{Five-fold cross-validation and evaluation.}
We employ five-fold cross-validation at the patient level, so that all acquisitions (TMAs) from the same patient---and hence all patches within those acquisitions---are assigned to the same fold and never simultaneously appear in both the training and validation splits, thereby preventing patient-level leakage. Class proportions are tracked across folds so that the per-fold class balance remains close to the cohort-wide balance. Classification performance is primarily assessed using the area under the precision--recall curve (AUPRC), and additionally reported using the area under the receiver operating characteristic curve (AUROC).

The ROC curve is defined by
\[
\mathrm{TPR}(t)=\frac{\mathrm{TP}(t)}{\mathrm{TP}(t)+\mathrm{FN}(t)},
\qquad
\mathrm{FPR}(t)=\frac{\mathrm{FP}(t)}{\mathrm{FP}(t)+\mathrm{TN}(t)},
\]
and AUROC is computed as
\[
\mathrm{AUROC}
=
\int_0^1 \mathrm{TPR}(\mathrm{FPR})\,d(\mathrm{FPR}).
\]
AUPRC is computed from the precision--recall curve:
\[
\mathrm{Precision}(t)=\frac{\mathrm{TP}(t)}{\mathrm{TP}(t)+\mathrm{FP}(t)},
\qquad
\mathrm{Recall}(t)=\frac{\mathrm{TP}(t)}{\mathrm{TP}(t)+\mathrm{FN}(t)},
\]
and
\[
\mathrm{AUPRC}
=
\int_0^1 \mathrm{Precision}(\mathrm{Recall})\,d(\mathrm{Recall}),
\]
which is approximated numerically using the standard average-precision estimator. Metrics are computed independently for each fold and summarized across folds.

\subsubsection{MIL survival analysis with Cox proportional hazards}

\paragraph{Risk prediction.}
For survival modeling, each slice $i$ is associated with an observed survival time $T_i\in\mathbb{R}_{+}$ and an event indicator $E_i\in\{0,1\}$, where $E_i=1$ denotes an observed event and $E_i=0$ indicates right censoring. A Cox prediction head $g_{\mathrm{cox}}(\cdot)$ maps the pooled representation to a scalar risk score:
\[
\hat{r}_i=g_{\mathrm{cox}}(\mathbf{s}_i)\in\mathbb{R}.
\]

\paragraph{Cox partial log-likelihood loss.}
Let the risk set be defined as $\mathcal{R}(i)=\{j:T_j\ge T_i\}$. The Cox partial log-likelihood is
\[
\ell_{\mathrm{cox}}
=
\sum_{i\in\mathcal{I}_{\mathrm{train}}}
E_i
\left(
\hat{r}_i
-
\log\sum_{j\in\mathcal{R}(i)}\exp(\hat{r}_j)
\right),
\]
and we minimize the normalized negative partial log-likelihood:
\[
\mathcal{L}_{\mathrm{cox}}
=
-\frac{1}{\sum_{i\in\mathcal{I}_{\mathrm{train}}}E_i}
\;\ell_{\mathrm{cox}}.
\]

\paragraph{Five-fold cross-validation and evaluation.}
Survival prediction is evaluated using five-fold cross-validation at the patient level, so that all acquisitions (TMAs) from the same patient are assigned to the same fold, preventing patient-level leakage between training and validation. Within each fold, the MIL--Cox model is trained on the training split and evaluated on the held-out split. Performance is quantified using the concordance index (C-index):
\[
\mathrm{C\text{-}index}
=
\mathbb{P}\bigl(\hat{r}_i>\hat{r}_j \,\big|\, T_i<T_j,\;E_i=1\bigr),
\]
which measures the fraction of correctly ordered comparable slice pairs.

For risk stratification, patients in each held-out fold are further stratified into high-risk and low-risk groups based on the median predicted risk $\hat{r}_i$ within that fold, and survival differences are visualized using Kaplan--Meier curves. Statistical separation between the two groups is assessed using a log-rank test.

\subsubsection{Zero-shot fusion retrieval--based biomarker inference} 
\label{sec:biomarker} 

\paragraph{Task formulation.} To evaluate whether tri-modal alignment enables improved biomarker inference through the integration of metadata-conditioned text and H\&E embeddings, we formulate biomarker inference as a zero-shot fusion retrieval task. Given a query patch with both H\&E and metadata-only text representations, we retrieve the most similar mIF patches from a held-out reference gallery and evaluate how faithfully the retrieved mIF biomarker patterns match the ground-truth mIF patch paired with the query. This task requires no task-specific training or fine-tuning; inference is performed entirely through retrieval in the pretrained Haiku embedding space.

\paragraph{Metadata-only text extraction.} To ensure that the text modality contributes only clinical and contextual information without directly encoding biomarker-level molecular profiles, we extract metadata-only text descriptions by removing all biomarker-specific content from the full text descriptions. Specifically, we apply pattern-based truncation to strip sentences following transition phrases that introduce molecular profile information (e.g., ``regarding the molecular profile'', ``in terms of protein expression''), retaining only the preceding tissue-level clinical context such as organ type, disease status, staging, and tissue annotation. This design ensures that any improvement in biomarker inference from fusion retrieval is attributable to complementary semantic knowledge rather than explicit biomarker supervision in the query text.

\paragraph{Fusion retrieval scoring.} For each query patch $k$ with H\&E embedding $\tilde{\mathbf{z}}^{\mathrm{HE}}_k$ and metadata-only text embedding $\tilde{\mathbf{z}}^{\mathrm{TXT}}_k$, the fused retrieval score against each candidate mIF embedding $\tilde{\mathbf{z}}^{\mathrm{mIF}}_j$ in the reference gallery is computed as
\[
s^{\mathrm{fusion}}_{k,j}
=
\alpha \left( \tilde{\mathbf{z}}^{\mathrm{HE}}_k \cdot \tilde{\mathbf{z}}^{\mathrm{mIF}}_j \right)
+
(1-\alpha) \left( \tilde{\mathbf{z}}^{\mathrm{TXT}}_k \cdot \tilde{\mathbf{z}}^{\mathrm{mIF}}_j \right)
=
\left( \alpha\,\tilde{\mathbf{z}}^{\mathrm{HE}}_k + (1-\alpha)\,\tilde{\mathbf{z}}^{\mathrm{TXT}}_k \right) \cdot \tilde{\mathbf{z}}^{\mathrm{mIF}}_j,
\]
where $\alpha \in [0,1]$ is a fixed fusion weight. The second equality shows that score-level fusion is equivalent to forming a single tri-modal fused query embedding $\alpha\,\tilde{\mathbf{z}}^{\mathrm{HE}}_k + (1-\alpha)\,\tilde{\mathbf{z}}^{\mathrm{TXT}}_k$ and retrieving against the mIF gallery in the shared embedding space; we use the same fused-query formulation in the counterfactual retrieval analyses below (Methods~\ref{sec:counterfactual_analysis}). The mIF candidate gallery, data splits, and spatial ground-truth definitions are identical to those used in unimodal cross-modality retrieval (Methods~\ref{sec:retrieval}), ensuring a fair and controlled comparison. We additionally evaluate unimodal retrieval baselines corresponding to $\alpha=1$ (H\&E-only) and $\alpha=0$ (Text-only). The fusion weight is optimized by grid search over $\alpha \in \{0, 0.1, 0.2, \ldots, 1.0\}$, selecting the value that maximizes mean Pearson correlation across all biomarkers on the held-out evaluation set. In our experiments, the optimal fusion weight is $\alpha = 0.8$.

\paragraph{Biomarker inference via Pearson correlation.} For each tissue region $r$ and biomarker channel $c \in \mathcal{C}_{\mathrm{valid}}$, we compute the mean biomarker abundance for each patch from the ground-truth mIF data. Let $\mu^{(r)}_{k,c}$ denote the mean intensity of biomarker $c$ in the ground-truth mIF patch at position $k$. For each retrieval strategy, the inferred biomarker abundance is obtained as a similarity-score-weighted sum across the top-$K$ retrieved mIF patches:
\[
\hat{\mu}^{(r)}_{k,c}
=
\frac{\sum_{j \in \mathcal{R}_k} s_{k,j} \cdot \mu_{j,c}}
{\sum_{j \in \mathcal{R}_k} s_{k,j}},
\]
where $\mathcal{R}_k$ denotes the set of top-$K$ retrieved mIF patches for query patch $k$ and $s_{k,j}$ is the fusion retrieval similarity score for candidate $j$. This weighted aggregation assigns higher influence to more confidently retrieved patches. Biomarker inference quality is then quantified by the Pearson correlation coefficient (PCC) computed across all patches within each region:
\[
\mathrm{PCC}^{(r)}_{c}
=
\mathrm{corr}\!\left(
\left\{\mu^{(r)}_{k,c}\right\}_{k=1}^{N_r},\;
\left\{\hat{\mu}^{(r)}_{k,c}\right\}_{k=1}^{N_r}
\right).
\]
This per-region, per-biomarker PCC captures how well the retrieval-based inference preserves the spatial pattern of biomarker expression across patches within each tissue region.

\paragraph{Global and per-biomarker aggregation.} Global PCC for each biomarker $c$ is computed by averaging $\mathrm{PCC}^{(r)}_{c}$ across all valid regions:
\[
\overline{\mathrm{PCC}}_c
=
\frac{1}{|\mathcal{R}_{\mathrm{valid}}|}
\sum_{r \in \mathcal{R}_{\mathrm{valid}}}
\mathrm{PCC}^{(r)}_{c}.
\]
The aggregate mean PCC across all biomarkers is reported as the primary summary metric. To ensure robustness, only biomarker channels present in at least $80\%$ of evaluation regions are retained, yielding a validated set $\mathcal{C}_{\mathrm{valid}}$ of 52 biomarkers.


\subsection{Counterfactual retrieval analysis and microenvironment stratification}
\label{sec:counterfactual_analysis}

This section defines the downstream analyses used to quantify counterfactual effects induced by metadata-only text-conditioned retrieval. Throughout, we use the notation and embedding definitions from the tri-modal retrieval framework. For each fixed H\&E query patch $x^{\mathrm{HE}}_i$ (with embedding $\tilde{\mathbf{z}}^{\mathrm{HE}}_i$), we construct two metadata-only text descriptions: a control description $x^{\mathrm{TXT},(0)}_i$ and a counterfactual description $x^{\mathrm{TXT},(1)}_i$, both containing clinical context (e.g., staging, diagnosis, tissue type, survival status) but excluding explicit biomarker-level molecular profiles. The corresponding embeddings $\tilde{\mathbf{z}}^{\mathrm{TXT},(0)}_i$ and $\tilde{\mathbf{z}}^{\mathrm{TXT},(1)}_i$ are obtained by encoding metadata-only descriptions through the pretrained text encoder. For $k\in\{0,1\}$, we form a fused query embedding as a convex combination,
\[
\tilde{\mathbf{z}}^{\mathrm{fusion},(k)}_{i}
=
\alpha\,\tilde{\mathbf{z}}^{\mathrm{HE}}_{i}
+
(1-\alpha)\,\tilde{\mathbf{z}}^{\mathrm{TXT},(k)}_{i},
\]
 We then retrieve the top-$K$ mIF patches from the held-out gallery using cosine similarity:
\[
s^{(k)}_{ij}
=
\tilde{\mathbf{z}}^{\mathrm{fusion},(k)}_{i}\cdot \tilde{\mathbf{z}}^{\mathrm{mIF}}_{j},
\qquad
\pi_i^{(k)}
=
\mathrm{argsort}_{j}(s^{(k)}_{ij}).
\]
The retrieved mIF index set under condition $k$ is defined as
\[
\mathcal{C}^{(k)}_i
=
\left\{\pi_i^{(k)}(1),\ldots,\pi_i^{(k)}(K)\right\}.
\]
All counterfactual analyses below compare the control retrieval set $\mathcal{C}^{(0)}_i$ against the counterfactual retrieval set $\mathcal{C}^{(1)}_i$, while keeping the H\&E query patch fixed.

\subsubsection{Metadata-only text construction for counterfactual analysis}
\label{sec:metadata_text_counterfactual}

To ensure that counterfactual predictions are driven by clinical context rather than explicit biomarker supervision, both control and counterfactual text descriptions are constructed as metadata-only descriptions. Specifically, we apply the same pattern-based truncation procedure described in Methods~\ref{sec:biomarker} to strip all biomarker-specific molecular profile information from the full text descriptions, retaining only tissue-level clinical metadata such as organ type, disease status, diagnosis, staging, grade, survival status, and tissue annotation. 

For the breast cancer counterfactual experiment, the control description specifies the original staging (T2N0M0, stage IIA, grade 2) while the counterfactual description modifies only the staging fields (T4N2M1, stage IV, grade 3), keeping all other metadata identical. For the lung cancer counterfactual experiment, the control description specifies the original survival status (Deceased, survival: 25 months) while the counterfactual description modifies only the survival fields (Alive, survival: 60 months), keeping all other clinical metadata including staging (T3N1M0, stage IIIA) identical.

Both control and counterfactual metadata-only text descriptions are encoded through the pretrained Haiku text encoder and broadcast to all query patches within the selected region, ensuring that all patches share the same text embedding under each condition. For counterfactual retrieval experiments, we use a fusion weight of $\alpha = 0.6$ (H\&E) and $1-\alpha = 0.4$ (text), which assigns a higher weight to the text modality compared with the biomarker inference setting ($\alpha = 0.8$) to amplify the effect of counterfactual text perturbations on the retrieval outcome.

\subsubsection{Patient-level subpopulation composition shift}
\label{sec:subpop_shift}

Each retrieved mIF patch $j$ is associated with a region identifier and corresponding clinical metadata. To quantify whether counterfactual retrieval changes the clinical composition of retrieved patches, we analyze TNM-derived categories associated with the retrieved region identifiers. Let $\ell(j)$ denote the TNM string associated with retrieved patch $j$ (when available). We parse $\ell(j)$ into its component categories and perform analyses independently for $T$ and $N$ stages.

For a category label set $\mathcal{Y}$ (e.g., $\mathcal{Y}=\{N0,N1,N2,\ldots\}$), we compute, for each query $i$, condition $k$, and category $c\in\mathcal{Y}$, the within-query proportion
\[
p_{i,k}(c)
=
\frac{1}{|\mathcal{C}^{(k)}_i|}
\sum_{j\in \mathcal{C}^{(k)}_i}
\mathbb{I}\!\left(\ell(j)=c\right).
\]
For each category $c$, we compare the paired distributions $\{p_{i,0}(c)\}_i$ and $\{p_{i,1}(c)\}_i$ using a paired two-sided Wilcoxon rank-sum (Mann--Whitney $U$) test across the $n$ matched queries. Multiple testing across categories is controlled using the Benjamini--Hochberg false discovery rate (FDR) procedure.

\subsubsection{H\&E embedding-based microenvironment clustering}
\label{sec:he_microenv_cluster}

To stratify query patches by morphological context, we perform unsupervised clustering on Haiku H\&E embeddings. For a selected tissue region, we collect the normalized H\&E embeddings $\{\tilde{\mathbf{z}}^{\mathrm{HE}}_i\}_{i=1}^{n}$ and apply $K$-means clustering with a fixed number of clusters $K_{\mathrm{clust}}=4$, yielding cluster assignments $c_i \in \{0,1,\ldots,K_{\mathrm{clust}}-1\}$ and cluster centroids $\boldsymbol{\mu}_k$. Each resulting cluster is then assigned a descriptive morphological compartment label (e.g., \emph{fibroblast-rich stroma}, \emph{epithelial-dominant tumor core}) by manual inspection of representative prototype patches (Methods~\ref{sec:prototype_selection}).

\subsubsection{Prototype patch selection for cluster interpretation}
\label{sec:prototype_selection}

To facilitate qualitative interpretation of each H\&E-derived cluster, we select representative prototype patches. For each cluster $k$, we compute Euclidean distances between cluster members and the cluster centroid in embedding space and select the $m$ closest patches, with $m=3$ when available:
\[
\mathcal{P}_k
=
\operatorname*{arg\,min}_{\substack{i:\,c_i=k \\ |\mathcal{P}_k|=m}}
\left\lVert \tilde{\mathbf{z}}^{\mathrm{HE}}_i - \boldsymbol{\mu}_k \right\rVert_2.
\]
The corresponding H\&E image crops are saved and used as visual prototypes for manual cluster annotation, in which each H\&E-derived cluster is assigned a descriptive morphological compartment name (e.g., \emph{fibroblast-rich stroma}, \emph{inflamed tumor zone}, \emph{epithelial-dominant tumor core}) by inspection of these prototype crops.

\subsubsection{Cluster-stratified biomarker differential testing under counterfactual retrieval}
\label{sec:cluster_biomarker_test}

To quantify counterfactual biomarker shifts under a fixed morphological context, we perform cluster-stratified analysis conditioned on H\&E-derived clusters. Specifically, for a given H\&E cluster $g$, we restrict attention to query patches whose embeddings are assigned to cluster $g$ and compare biomarker abundance patterns between the original and counterfactual retrieval results.

Let $\mathcal{I}_g = \{ i \mid c_i = g \}$ denote the set of query patches assigned to H\&E cluster $g$. For each query patch $i \in \mathcal{I}_g$, metadata-only fusion-based retrieval yields two mIF patch sets:
\[
\mathcal{C}^{(0)}_i \quad \text{(original condition)}, 
\qquad
\mathcal{C}^{(1)}_i \quad \text{(counterfactual condition)}.
\]

\paragraph{Weighted mean biomarker abundance summarization.}
For each retrieved mIF patch $j$ and biomarker channel $c \in \mathcal{C}_{\mathrm{valid}}$, we obtain the mean biomarker abundance $\mu_{j,c}$ from the associated spatial biomarker quantification. For each query patch $i$, condition $k \in \{0,1\}$, and biomarker $c$, we summarize biomarker abundance over the top-$K$ retrieved mIF patches using similarity-score-weighted averaging:
\[
\bar{\mu}^{(k)}_{i,c}
=
\frac{\sum_{j \in \mathcal{C}^{(k)}_i} s^{(k)}_{ij} \cdot \mu_{j,c}}
{\sum_{j \in \mathcal{C}^{(k)}_i} s^{(k)}_{ij}},
\]
where $s^{(k)}_{ij}$ denotes the fusion retrieval similarity score for patch $j$ under condition $k$. This score-weighted formulation assigns higher influence to more confidently retrieved patches.

\paragraph{Counterfactual abundance shift.}
For each query patch $i \in \mathcal{I}_g$ and biomarker $c$, we define the counterfactual abundance shift as
\[
d_{i,c}
=
\bar{\mu}^{(1)}_{i,c}
-
\bar{\mu}^{(0)}_{i,c}.
\]
Positive values of $d_{i,c}$ indicate increased biomarker abundance under the counterfactual condition, while negative values indicate decreased abundance.

\paragraph{Statistical testing within clusters.}
For each biomarker $c \in \mathcal{C}_{\mathrm{valid}}$, we test whether the distribution
\[
\{ d_{i,c} \}_{i \in \mathcal{I}_g}
\]
differs significantly from zero using a two-sided Wilcoxon signed-rank test, which accounts for paired comparisons at the query level. To control for multiple testing across biomarkers, p-values are adjusted using the Benjamini--Hochberg false discovery rate (FDR) procedure.

This cluster-stratified, abundance-based testing framework isolates molecular shifts attributable to counterfactual metadata-only semantic intervention under fixed H\&E morphology, while the score-weighted averaging ensures that retrieval confidence is appropriately accounted for in the biomarker summarization.

\subsubsection{PCA of per-patch counterfactual biomarker shift profiles}
\label{sec:pca_shift}

To quantify heterogeneity of counterfactual biomarker abundance changes within a fixed microenvironment cluster, we assemble the per-patch shift vectors
\[
\Delta \mathbf{b}_i
=
\left(d_{i,1},d_{i,2},\ldots,d_{i,C}\right)\in\mathbb{R}^{C},
\qquad i\in\mathcal{I}_g,
\]
into a shift matrix
\[
\Delta \mathbf{B}
=
\left[\Delta \mathbf{b}_i\right]_{i\in\mathcal{I}_g}
\in
\mathbb{R}^{|\mathcal{I}_g|\times C}.
\]
Biomarker channels with missing values for all patches are removed. To enable PCA without imputation, patches with incomplete shift vectors after channel filtering are excluded. We then perform PCA on $\Delta \mathbf{B}$ and obtain 2D coordinates for each patch:
\[
\mathbf{u}_i
=
\mathrm{PCA}(\Delta \mathbf{b}_i)
\in\mathbb{R}^{2},
\qquad
\mathbf{u}_i=(u_{i,1},u_{i,2}).
\]
The resulting PCA projection summarizes dominant modes of variation in counterfactual biomarker shifts across patches.

\subsubsection{Association between baseline biomarker state and counterfactual shift trajectories}
\label{sec:baseline_pc_assoc}

Finally, we test whether baseline biomarker state is associated with the dominant axes of counterfactual change. For each query patch $i\in\mathcal{I}_g$, we compute the baseline biomarker vector from the original matched mIF patch $x^{\mathrm{mIF}}_{\mathrm{orig},i}$ (i.e., the mIF patch aligned to the fixed H\&E query prior to retrieval). Baseline biomarker means are computed as
\[
m^{\mathrm{orig}}_{i,c}
=
\frac{1}{P^2}\sum_{h=1}^{P}\sum_{w=1}^{P} x^{\mathrm{mIF}}_{\mathrm{orig},i}(h,w,c),
\qquad c=1,\ldots,C.
\]
We then compute Pearson correlation coefficients between $\{m^{\mathrm{orig}}_{i,c}\}_{i\in\mathcal{I}_g}$ and the PCA coordinates $\{u_{i,1}\}_{i\in\mathcal{I}_g}$ and $\{u_{i,2}\}_{i\in\mathcal{I}_g}$:
\[
\rho_{c,1}
=
\mathrm{corr}\!\left(\{m^{\mathrm{orig}}_{i,c}\}_{i\in\mathcal{I}_g},\{u_{i,1}\}_{i\in\mathcal{I}_g}\right),
\qquad
\rho_{c,2}
=
\mathrm{corr}\!\left(\{m^{\mathrm{orig}}_{i,c}\}_{i\in\mathcal{I}_g},\{u_{i,2}\}_{i\in\mathcal{I}_g}\right).
\]
These correlations are summarized as a biomarker-by-PC correlation matrix.

\section*{Author Contributions}

Y.C., Z.H., J.L., Z.W., A.E.T. conceived the study. Y.C., J.L., Z.H., Z.W., A.E.T. developed the methodology. Y.C., J.L. performed the experiments. W.L., D.K., Y.D., A.M. provided essential feedback. Y.C., Z.H., Z.W., A.E.T. wrote the manuscript. Z.H., A.E.T., Z.W. supervised the study. All authors discussed the results and approved the final manuscript.

\section*{Acknowledgements}
This project was supported by the startup funding from the Perelman School of Medicine, University of Pennsylvania (Z.H.). We also thank the authors of VirTues~\cite{Wenckstern2025-ve} for open-sourcing their code and pretraining framework, on which our mIF encoder pretraining was based.

\section*{Data Availability}

Demo data is available on Hugging Face at \url{https://huggingface.co/datasets/zhihuanglab/Haiku-demo-data}.

\section*{Code Availability}

Code is available at \url{https://github.com/zhihuanglab/Haiku}. Model checkpoint is available at \url{https://huggingface.co/zhihuanglab/Haiku}.

\section*{Conflict of Interests}
Aaron T. Mayer, Zhenqin Wu, and Alexandro E. Trevino are employees of Enable Medicine, Inc.

\clearpage
\printbibliography
\clearpage
\appendix
\setcounter{figure}{0}
\renewcommand{\thefigure}{S\arabic{figure}}

\begin{figure}[hbtp]
    \centering
    \includegraphics[width=\textwidth]{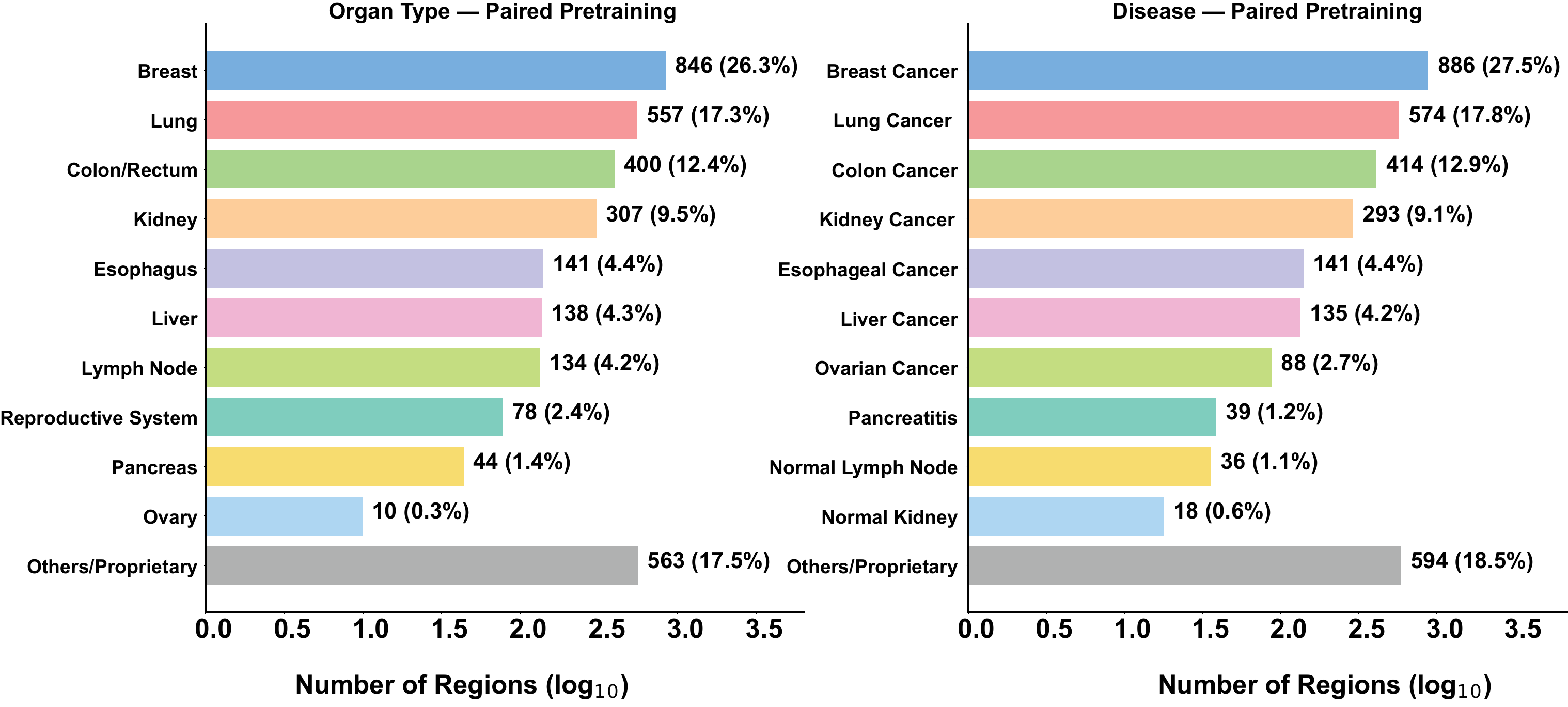}
    \caption{%
    \small \textbf{Composition of the paired pretraining set by organ type and disease.}
    Distribution of the 3{,}218 paired tissue slices used for Haiku tri-modal contrastive pretraining. Left, distribution by organ type, led by breast ($846$; $26.3\%$), lung ($557$; $17.3\%$), colon/rectum ($400$; $12.4\%$), and kidney ($307$; $9.5\%$), with additional contributions from esophagus, liver, lymph node, reproductive system, pancreas, and ovary, together with an aggregated others/proprietary category ($563$; $17.5\%$) covering remaining organs. Right, distribution by disease annotation, led by breast cancer ($886$; $27.5\%$), lung cancer ($574$; $17.8\%$), colon cancer ($414$; $12.9\%$), and kidney cancer ($293$; $9.1\%$), alongside esophageal, liver, and ovarian cancers, pancreatitis, normal lymph node and kidney samples, together with an aggregated others/proprietary category ($594$; $18.5\%$). The $x$-axis shows the number of regions on a $\log_{10}$ scale; numbers in parentheses indicate the percentage of total slices in the paired pretraining set.
    }
    \label{supplementary:1}
\end{figure}

\begin{figure}[hbtp]
    \centering
    \includegraphics[width=0.6\textwidth]{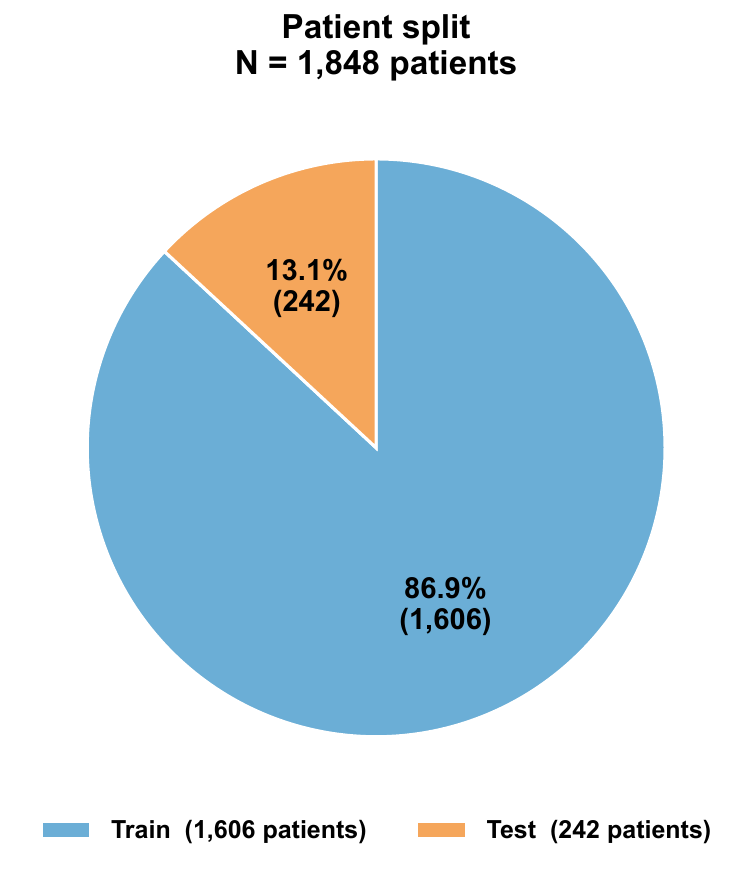}
    \caption{%
    \small \textbf{Patient-level train/test split.}
    Patient-level partition of the full Haiku cohort into training and held-out test sets. Of $N = 1{,}848$ patients in total, $1{,}606$ ($86.9\%$) are assigned to the training pool and $242$ ($13.1\%$) are reserved for the held-out test set. The split is performed at the patient level so that all tissue slices and patches originating from a given patient remain within a single partition, preventing patient-level leakage between training and evaluation across all downstream protocols.
    }
    \label{supplementary:2}
\end{figure}

\begin{figure}[hbtp]
    \centering
    \includegraphics[width=\textwidth]{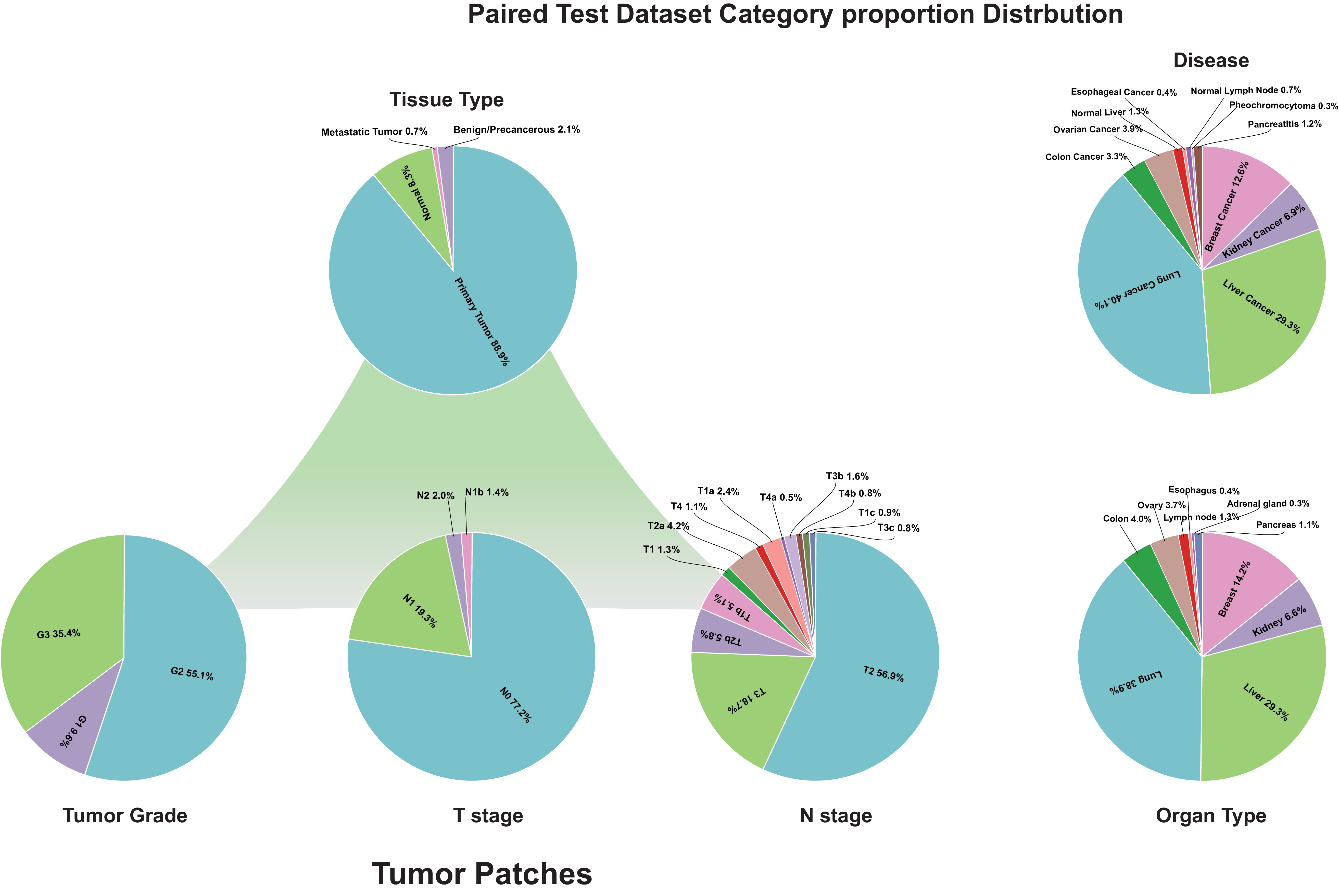}
    \caption{%
    \small \textbf{Paired test dataset: tumor-patch label composition across six clinically relevant categories.}
    Category proportions for the 336-slice paired held-out test set. \textbf{Tissue Type}, \textbf{Organ Type}, and \textbf{Disease} are three parallel dataset-level category axes computed across all 336 slices: \textbf{Tissue Type}: primary tumor ($88.9\%$), normal ($8.3\%$), benign/precancerous ($2.1\%$), and metastatic tumor ($0.7\%$); \textbf{Organ Type}: lung ($38.9\%$), liver ($29.3\%$), breast ($14.2\%$), kidney ($6.6\%$), colon ($4.0\%$), ovary ($3.7\%$), lymph node ($1.3\%$), pancreas ($1.1\%$), esophagus ($0.4\%$), and adrenal gland ($0.3\%$); \textbf{Disease}: lung cancer ($40.1\%$), liver cancer ($29.3\%$), breast cancer ($12.6\%$), kidney cancer ($6.9\%$), ovarian cancer ($3.9\%$), colon cancer ($3.3\%$), normal liver ($1.3\%$), pancreatitis ($1.2\%$), normal lymph node ($0.7\%$), esophageal cancer ($0.4\%$), and pheochromocytoma ($0.3\%$). The remaining three categories --- \textbf{N Stage}, \textbf{T Stage}, and \textbf{Tumor Grade} --- are sub-labels of the tumor-patch subset of Tissue Type (i.e., primary tumor and metastatic tumor patches, $89.6\%$ of the test set) and are defined only for tumor tissue: \textbf{N Stage (TNM)}: N0 ($77.2\%$), N1 ($19.3\%$), N2 ($2.0\%$), and N1b ($1.4\%$); \textbf{T Stage (TNM)}: T2 ($56.9\%$), T3 ($18.7\%$), T2b ($5.8\%$), T1b ($5.1\%$), T2a ($4.2\%$), T1a ($2.4\%$), T3b ($1.6\%$), T1 ($1.3\%$), T4 ($1.1\%$), T1c ($0.9\%$), T3c ($0.8\%$), T4b ($0.8\%$), and T4a ($0.5\%$); \textbf{Tumor Grade}: G2 ($55.1\%$), G3 ($35.4\%$), and G1 ($9.6\%$). Together, these distributions illustrate the substantial class imbalance that the linear-probing and zero-shot classification protocols must accommodate while retaining clinically meaningful coverage across organs, diseases, stages, and grades.    }
    \label{supplementary:3}
\end{figure}

\begin{figure}[hbtp]
    \centering
    \includegraphics[width=0.85\textwidth]{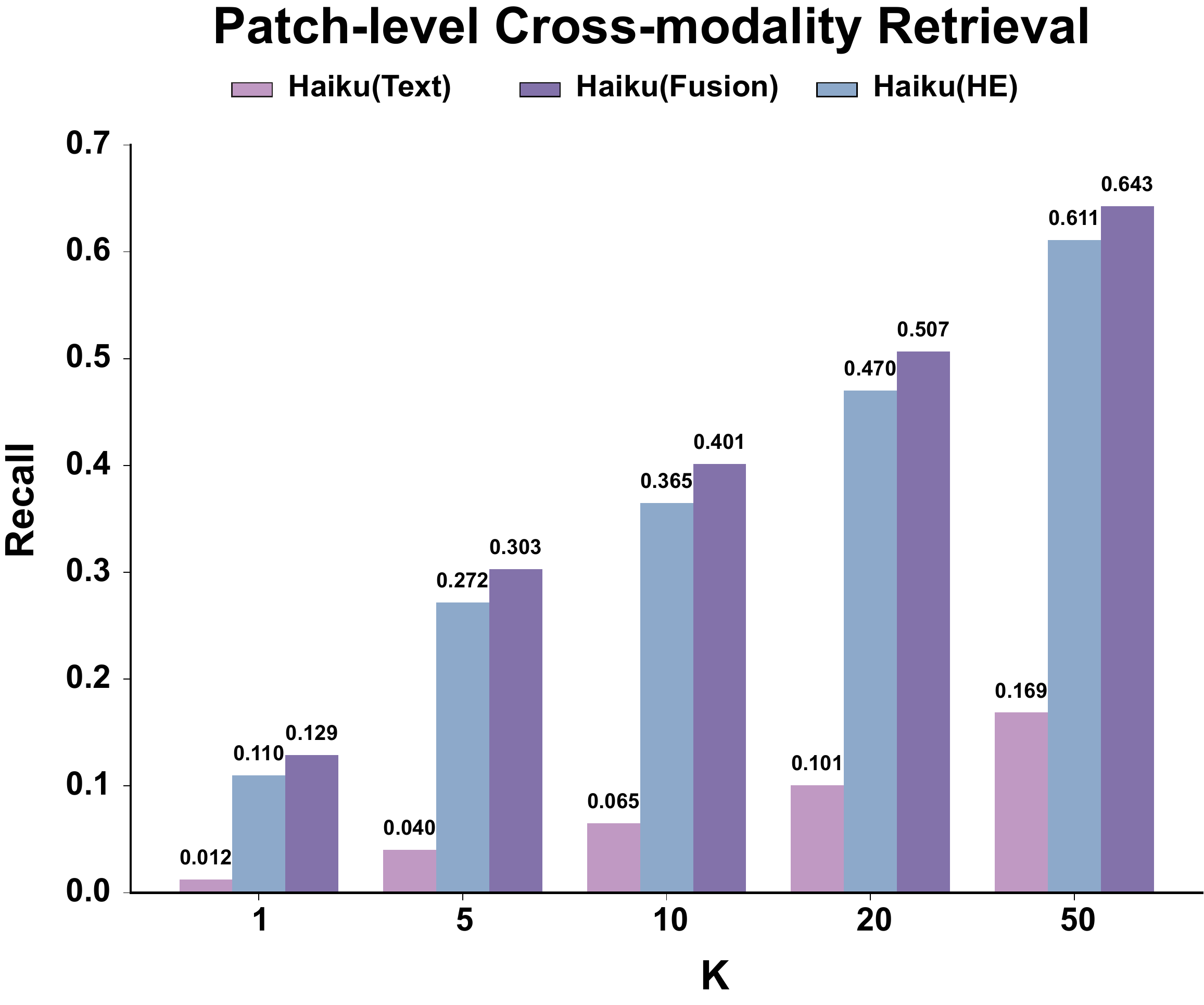}
    \caption{%
    \small \textbf{Patch-level cross-modality retrieval to mIF across query strategies.}
    Recall@$K$ (with $K \in \{1, 5, 10, 20, 50\}$) for patch-level cross-modality retrieval against the held-out mIF reference atlas, evaluated under three query strategies that all share mIF as the target modality: Haiku(Text), text-only query (metadata-only text descriptions); Haiku(H\&E), H\&E-only query; and Haiku(Fusion), the weighted-fusion query combining H\&E and text similarity scores ($\alpha = 0.8$ for H\&E, $1-\alpha = 0.2$ for text; Methods~\ref{sec:biomarker}). At Recall@$50$, Haiku(Text) reaches $0.169$, Haiku(H\&E) reaches $0.611$ (consistent with the H\&E-to-mIF retrieval result reported in \textbf{Figure~\ref{fig:figure2}e}), and Haiku(Fusion) reaches $0.643$, demonstrating that combining H\&E and metadata text outperforms either unimodal query alone for retrieval into the mIF atlas. The improvement of fusion over H\&E-only is consistent across all values of $K$.
    }
    \label{supplementary:4}
\end{figure}

\begin{figure}[hbtp]
    \centering
    \includegraphics[width=0.75\textwidth]{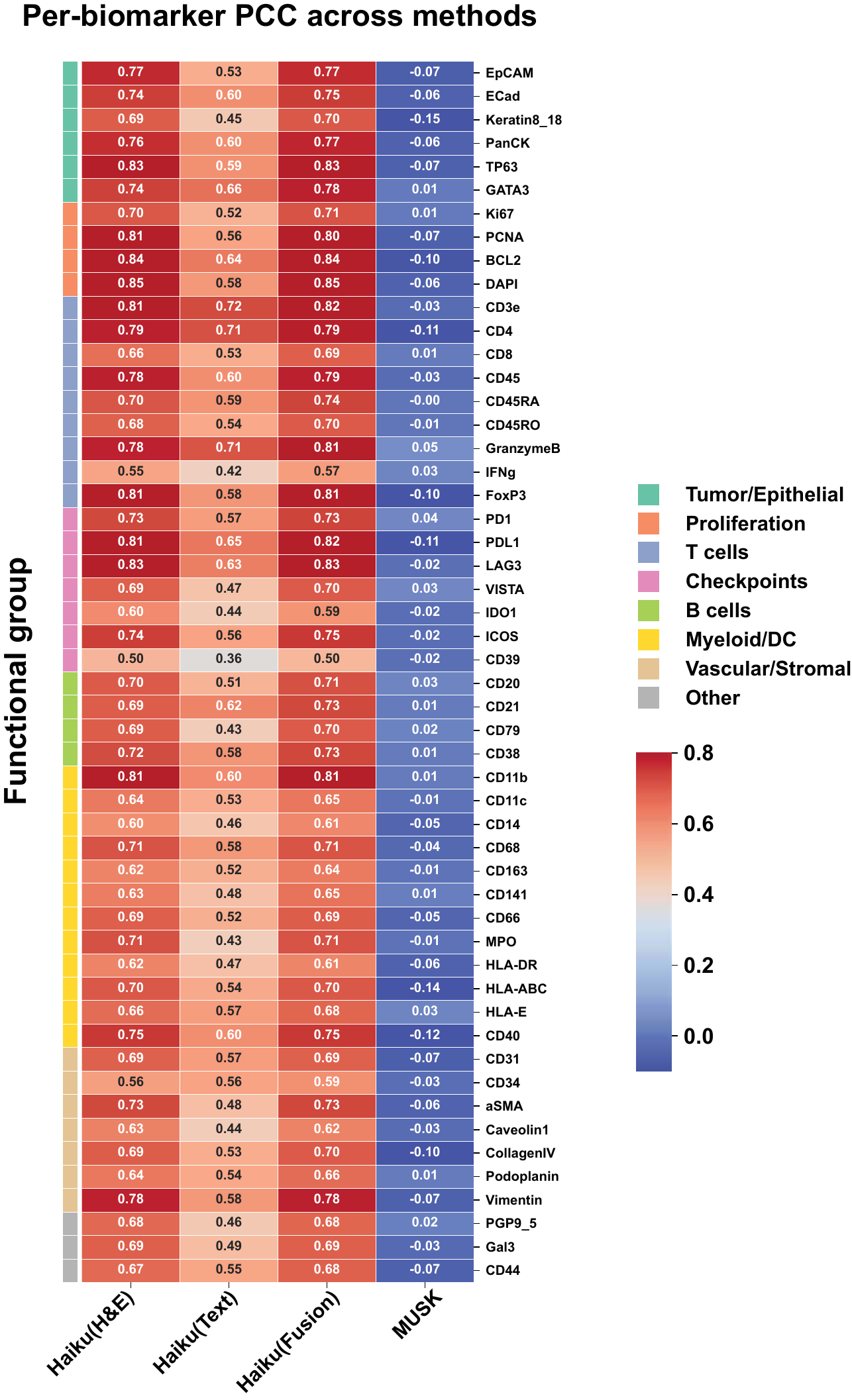}
    \captionsetup{
      format=plain,
      justification=raggedright,
      singlelinecheck=false,
      margin=0pt,
      width=\dimexpr\textwidth\relax
    }
    \caption[\textbf{Per-biomarker Pearson correlation coefficients (PCC) for biomarker inference across methods.}]%
    {\textbf{Per-biomarker Pearson correlation coefficients (PCC) for biomarker inference across methods.} See next page for caption.}
    \label{supplementary:5}
\end{figure}

\begin{figure}[hbtp]
    \ContinuedFloat
    \captionsetup{list=no}
    \caption{%
    \small \textbf{Per-biomarker Pearson correlation coefficients (PCC) for biomarker inference across methods.}
    Complete table of per-biomarker mean PCC values between retrieval-based predicted biomarker abundance and ground-truth mIF biomarker abundance, aggregated across the 336 paired held-out regions and reported for each of the 52 validated biomarker channels. Rows are organized by biological program (proliferation/cell cycle, immune activation, cytotoxic/effector, T-cell lineage, immune exhaustion, B-cell lineage, neural/other, myeloid/innate, stromal/vascular/ECM, nuclear/DNA, epithelial/differentiation, and survival/anti-apoptotic), matching the grouping used in Figure~\ref{fig:figure4}c--g. Columns compare the four retrieval strategies: Haiku(H\&E) using H\&E embeddings alone; Haiku(Text) using metadata-only text embeddings; Haiku(Fusion) combining the two with optimized weights ($0.8$ H\&E $+$ $0.2$ Text); and the MUSK baseline. Column-wise means are reported at the bottom: Haiku(H\&E) $0.710$, Haiku(Text) $0.547$, Haiku(Fusion) $0.718$, and MUSK $-0.033$. This table complements the box-plot summaries in Figure~\ref{fig:figure4}c--g by reporting the exact numerical value for each biomarker--method combination, enabling direct inspection of method-wise differences at the single-biomarker level.
    }
\end{figure}

\end{document}